\documentclass[final,3p,times,twocolumn]{elsarticle}

\usepackage{amssymb}

\usepackage{multirow}
\usepackage{hyperref}
\usepackage{times}
\usepackage{epsfig}
\usepackage{graphicx}
\usepackage{amsmath}
\usepackage{amssymb}
\usepackage[ruled,lined,linesnumbered]{algorithm2e}
\newcommand{\bs}[1]{\boldsymbol{#1}}
\newcommand{\transpose}{t}
\newcommand{\transp}{\transpose}

\usepackage{xcolor}
\usepackage{hyperref}
\usepackage{fancyhdr}
\hypersetup{%
  colorlinks=false,
  urlbordercolor=blue,
  pdfborderstyle={/S/U/W 1}
}

\journal{Computerized Medical Imaging and Graphics}

\begin{document}

\begin{frontmatter}
	
\title{Computational Pathology: Challenges and Promises for Tissue Analysis}
	
\author[label1,label2]{Thomas J. Fuchs}
\author[label1,label2]{Joachim M. Buhmann}
\address[label1]{Department of Computer Science, ETH Zurich,
         Universitaetstrasse 6, CH-8092 Zurich, Switzerland} 
\address[label2]{Competence Center for Systems Physiology and
         Metabolic Diseases, ETH Zurich, Schafmattstr. 18, CH-8093
         Zurich, Switzerland} 
	
\begin{abstract}
  The histological assessment of human tissue has emerged as the key
  challenge for detection and treatment of cancer. A plethora of
  different data sources ranging from tissue microarray data to gene
  expression, proteomics or metabolomics data provide a detailed
  overview of the health status of a patient. Medical
  doctors need to assess these information sources and they rely on
  data driven automatic analysis tools. Methods for classification,
  grouping and segmentation of heterogeneous data sources as well as
  regression of noisy dependencies and estimation of survival
  probabilities enter the processing workflow of a pathology diagnosis
  system at various stages. This paper reports on state-of-the-art of
  the design and effectiveness of computational pathology workflows
  and it discusses future research directions in this emergent field
  of medical informatics and diagnostic machine learning.
\end{abstract}


\end{frontmatter}
	

\footnotesize
\setcounter{tocdepth}{1}
\tableofcontents
\normalsize

\section{Computational Pathology: The systems view} 
  \label{sec:comp.path.def} 	
\thispagestyle{fancy}	
Modern pathology studies of biopsy tissue encompass multiple stainings
of histological material, genomics and proteomics analyses as well as
comparative statistical analyses of patient data. Pathology lays
not only a scientific foundation for clinical medicine but also serves
as a bridge between the fundamental sciences in natural science,
medicine and patient care.  Therefore, it can be viewed as one of the
key hubs for translational research in the health and life sciences,
subsequently facilitating translational medicine. In particular, the
abundance of heterogeneous data sources with a substantial amount of
randomness and noise poses challenging problems for statistics and
machine learning. Automatic processing of this wealth of data promises
a standardized and hopefully more objective diagnosis of the disease
state of a patient than manual inspection can provide today. An
automatic computational pathology pipeline also enables the medical
user to quantitatively benchmark the processing pipeline and to
identify error sensitive processing steps which can substantially
degrade the final prediction of survival times.

\subsection{Definition}\label{ssec:definition}

Computational Pathology as well as the medical discipline Pathology is
a wide and diverse field which encompass scientific research
as well as day-to-day work in medical clinics. The following definition
is an attempt for a concise and practical description of this novel
field:
	
\begin{quote}
  Computational Pathology investigates a complete probabilistic
  treatment of scientific and clinical workflows in general pathology,
  i.e. it combines experimental design, statistical pattern
  recognition and survival analysis to an unified framework for
  answering scientific and clinical questions in pathology.
\end{quote}

\begin{figure*}[htbp]
  \begin{center}					
    \includegraphics[width=1\linewidth]{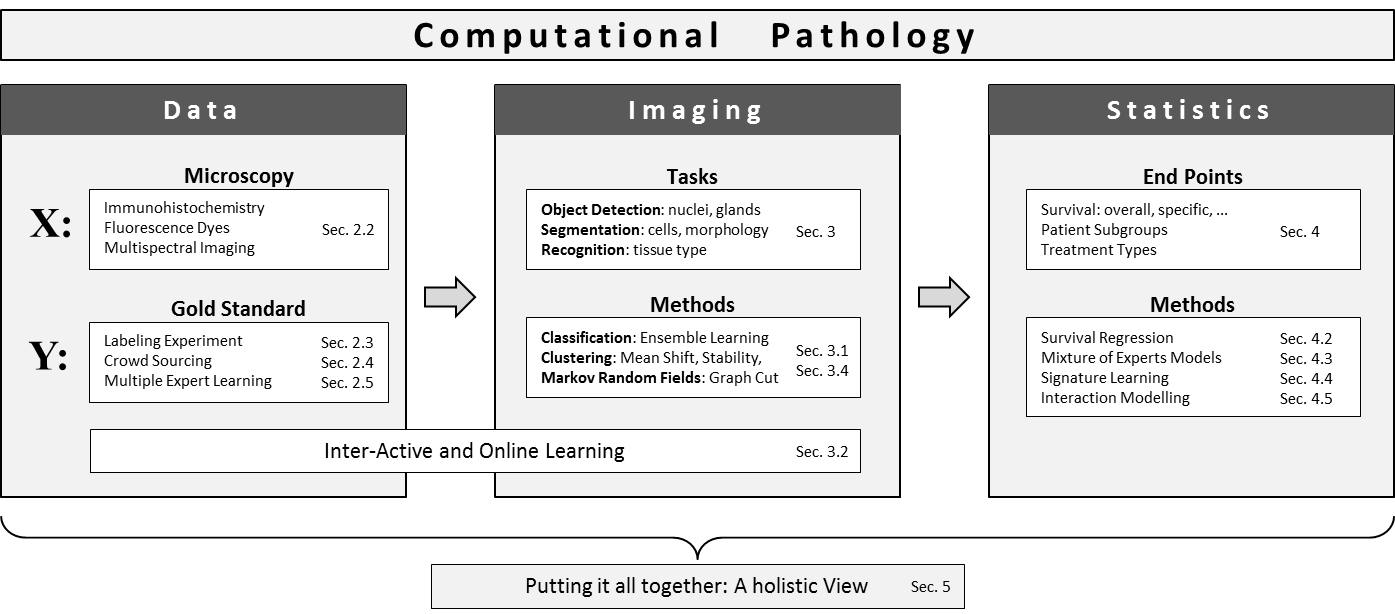}
  \end{center}			
  \caption{Schematic overview of the workflow in computational pathology 
    		  comprising of three major parts: 
    		  (i) the covariate data $X$ is acquired via 
    		  microscopy and the target data $Y$ is generated in
    		  extensive labeling experiments;
    		  (ii) image analysis in terms of nuclei detection,
    		  cell segmentation or texture classification 
    		  provides detailed information about the tissue;
  			  (iii) medical statistics, i.e. survival regression
  			  and mixture of expert models are used to investigate
  			  the clinical end point of interesting using data
  			  from the previous two parts. 
  			  The aim is to build a complete probabilistic
  			  workflow comprising all three parts.
  			  (The corresponding sections of the paper are noted 
  			  in addition.)
    \label{fig:comppath}
  }
\end{figure*}

Figure \ref{fig:comppath} depicts a schematic overview of the
field and the three major parts it consists of: data generation,
image analysis and medical statistics.

\section{Data: Tissue and Ground Truth}\label{sec:ground.truth}		
	
\subsection{Clear Cell Renal Cell Carcinoma}\label{subsec:introRCC}
Throughout this review we use Renal cell carcinoma (RCC) as a 
disease case to design and optimize a computational pathology framework
because it exhibits a number of properties which are highly relevant
for computational pathology. 
			
Renal cell carcinoma figures as one of the ten most frequent
malignancies in the statistics of Western societies
\cite{Grignon04}. The prognosis of renal cancer is poor since many
patients suffer already from metastases at the time of first
diagnosis. The identification of biomarkers for prediction of
prognosis (prognostic marker) or response to therapy (predictive
marker) is therefore of utmost importance to improve patient prognosis
\cite{Tannapfel96}. Various prognostic markers have been suggested in
the past \cite{Moch99, Sudarshan06}, but estimates of conventional
morphological parameters still provide most valuable information for
therapeutical decisions.
			
Clear cell RCC (ccRCC) emerged as the most common subtype of renal
cancer and it is composed of cells with clear cytoplasm and typical
vessel architecture. ccRCC exhibits an architecturally diverse
histological structure, with solid, alveolar and acinar patterns. The
carcinomas typically contain a regular network of small thin-walled
blood vessels, a diagnostically helpful characteristic of this
tumor. Most ccRCC show areas with hemorrhage or necrosis
(Fig. \ref{fig:rcc}d), whereas an inflammatory response is
infrequently observed.  Nuclei tend to be round and uniform with
finely granular and evenly distributed chromatin. Depending upon the
grade of malignancy, nucleoli may be inconspicuous and small, or large
and prominent, with possibly very large nuclei or bizarre nuclei
occurring \cite{Grignon04}.
					
The prognosis for patients with RCC depends mainly on the pathological
stage and the grade of the tumor at the time of surgery. Other
prognostic parameters include proliferation rate of tumor cells and
different gene expression patterns. Tannapfel et
al. \cite{Tannapfel96} have shown that cellular proliferation
potentially serves as another measure for predicting biological
aggressiveness and, therefore, for estimating the
prognosis. Immunohistochemical assessment of the MIB-1 (Ki-67) antigen
indicates that MIB-1 immunostaining (Fig. \ref{fig:rcc}d) is an
additional prognostic parameter for patient outcome. TMAs were highly
representative of proliferation index and histological grade using
bladder cancer tissue \cite{Nocito01}.
					
The TNM staging system specifies the local extension of the primary tumour
(T), the involvement of regional lymph nodes (N), and the presence of distant
metastases (M) as indicators of the disease state.
\cite{Wild09} focuses on reassessing the current TNM staging system
for RCC and concludes that outcome prediction for RCC remains
controversial. Although many parameters have been tested for
prognostic significance, only a few have achieved general acceptance
in clinical practice. An especially interesting observation of
\cite{Wild09} is that multivariate Cox proportional hazards regression
models including multiple clinical and pathologic covariates were more
accurate in predicting patient outcome than the TNM staging system.
On one hand this finding demonstrates the substantial difficulty of the
task and on the other hand it is a motivation for research in
computational pathology to develop robust machine learning frameworks
for reliable and objective prediction of disease progression.

\subsection{Tissue Microarrays}
			
\begin{figure*}[tb]
  \begin{center}
    \begin{tabular}{ccccc}
	\includegraphics[height=1.46in] {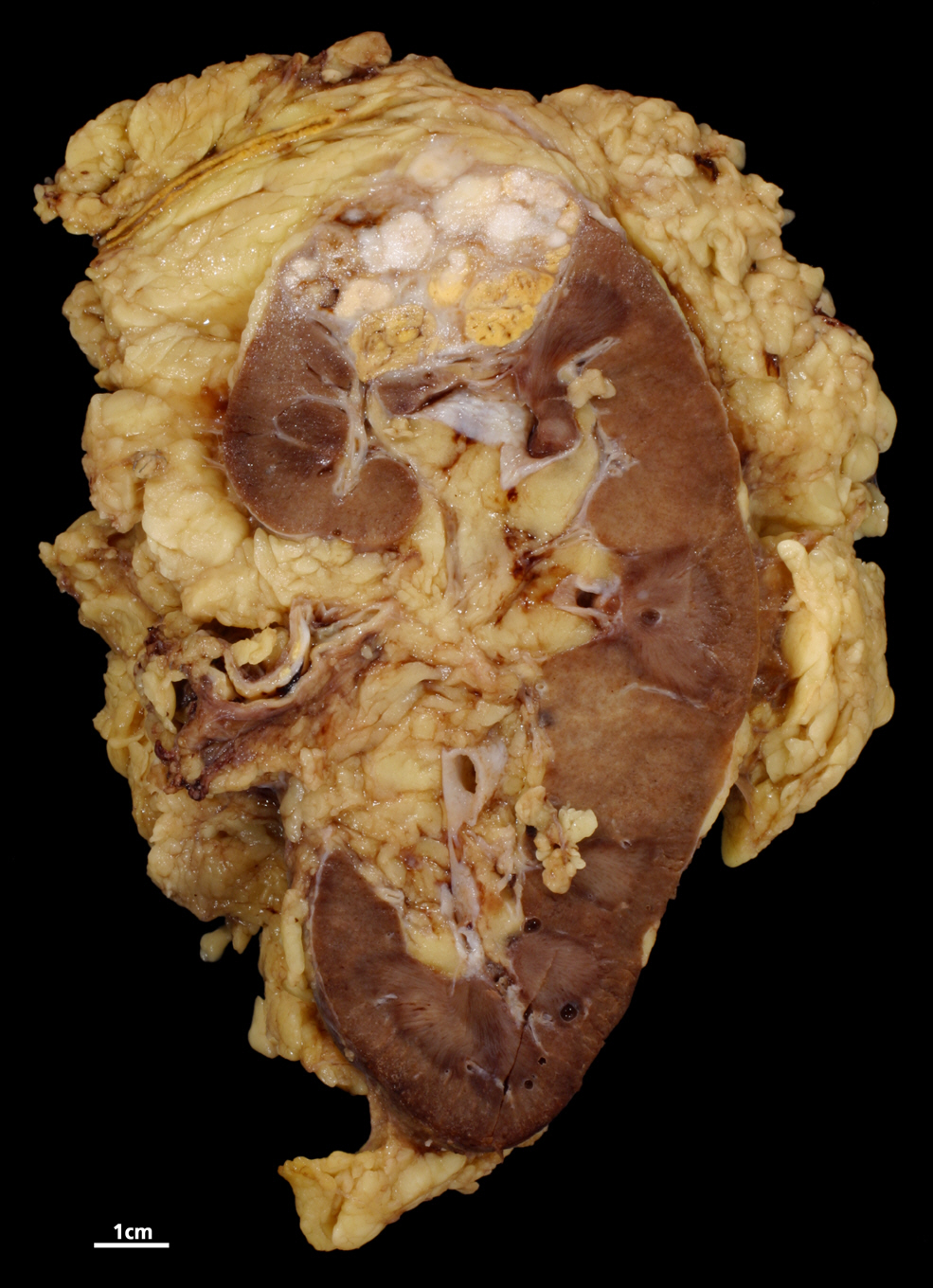} &
	  \includegraphics[height=1.46in] {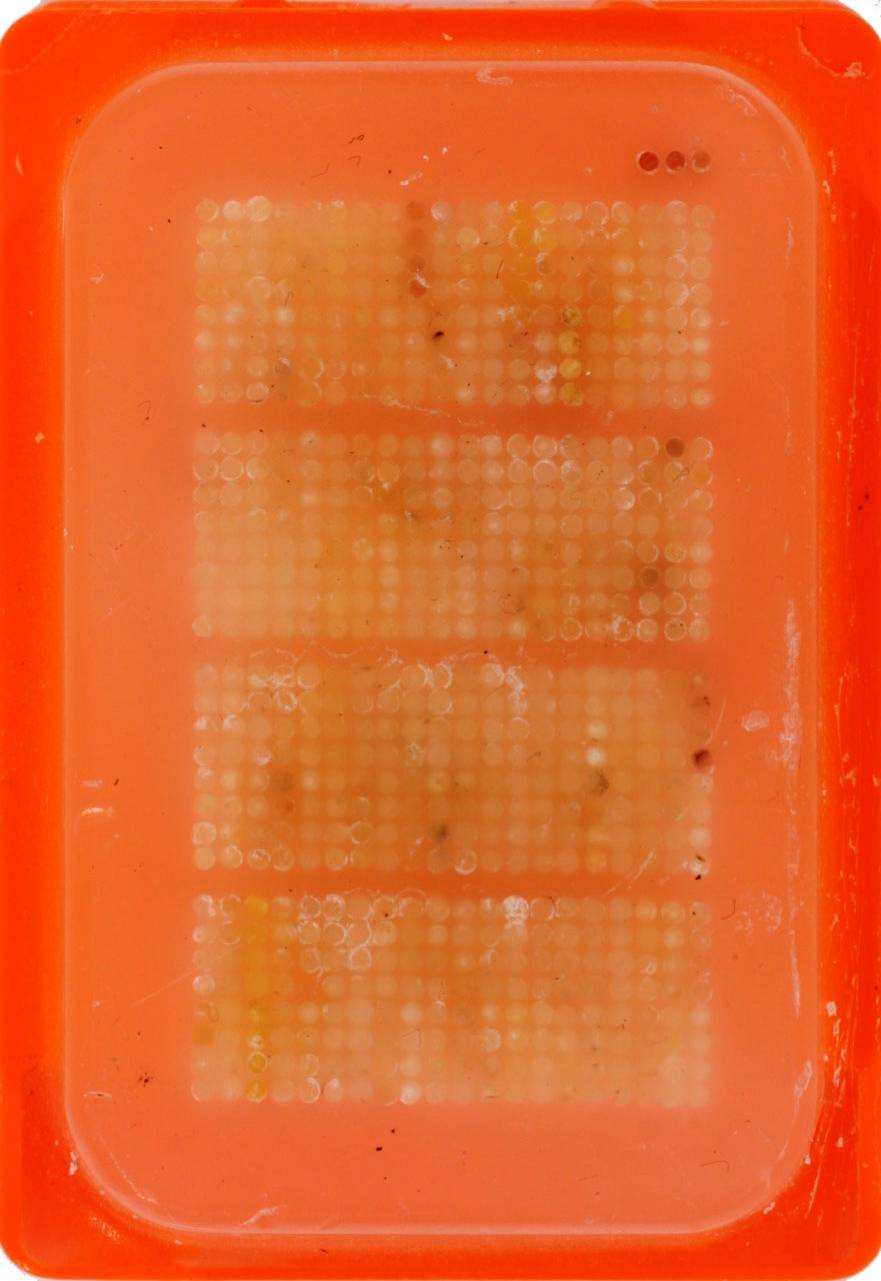} &
	  \includegraphics[height=1.46in] {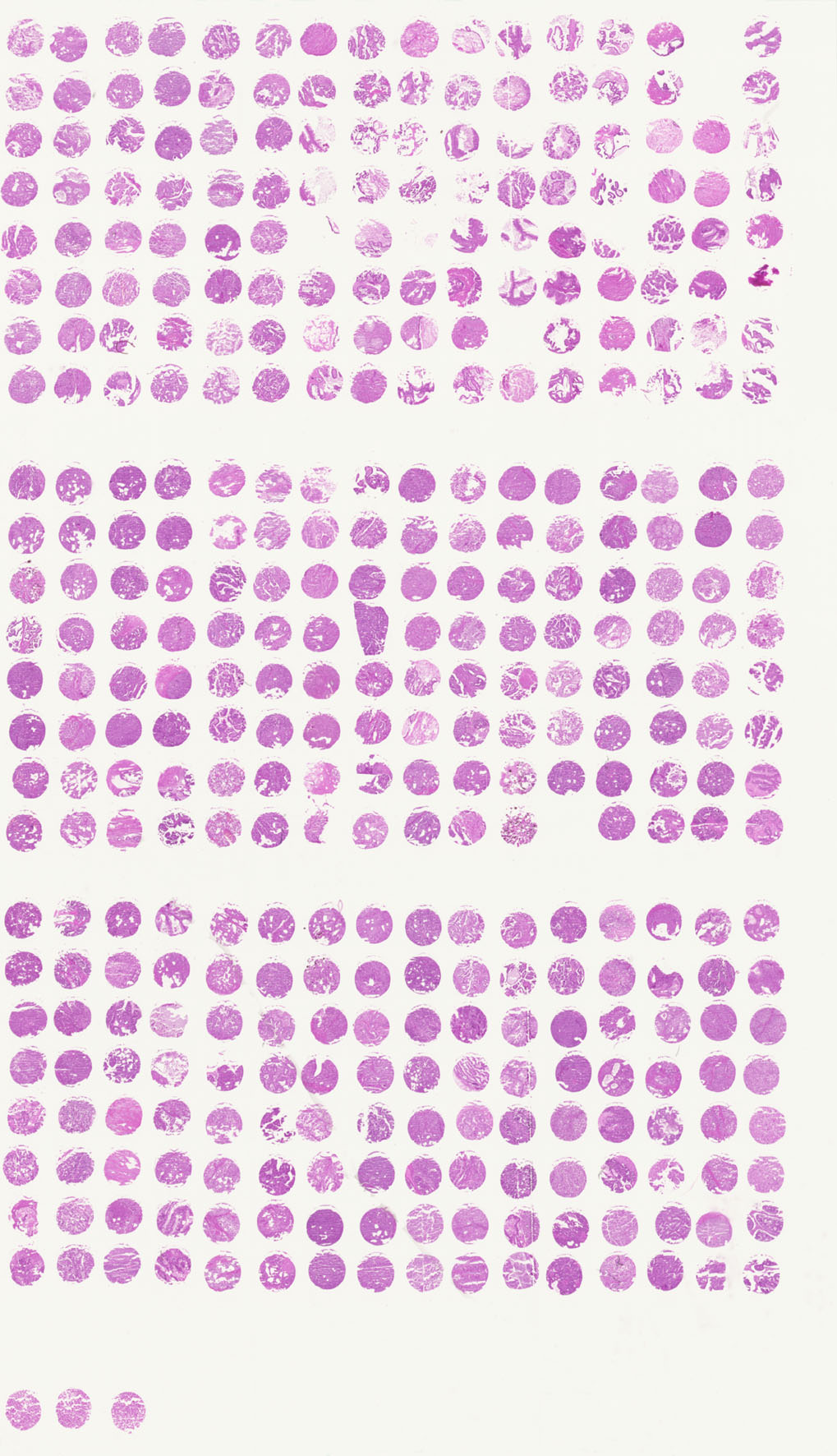} &	                           
	\includegraphics[height=1.46in] {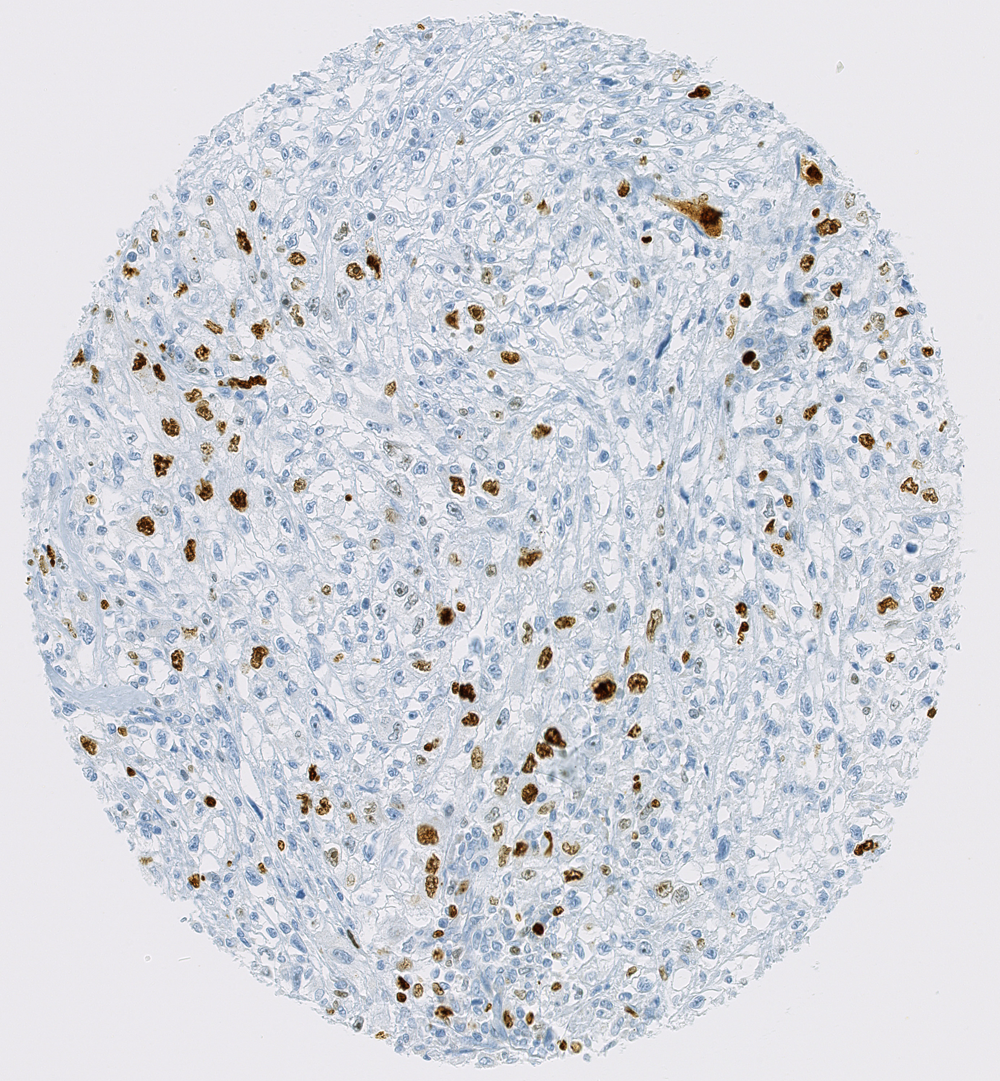} &
	\includegraphics[height=1.46in] {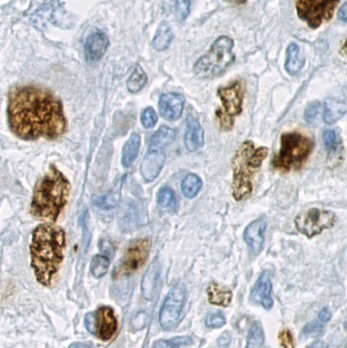} \\
	  (a) & (b) & (c) & (d) & (e)\\
     \end{tabular}
     \caption{ Tissue Microarray Analysis (TMA): Primary tissue
       samples are taken from a cancerous kidney (a).  Then $0.6 mm$
       tissue cylinders are extracted from the primary tumor material of
       different patients and arrayed in a recipient paraffin block
       (b).  Slices of $0.6\mu m$ are cut off the paraffin block and
       are immunohistochemically stained (c).  These slices are
       scanned and each spot, represents a different patient. Image
       (d) depicts a TMA spot of clear cell renal cell carcinoma
       stained with MIB-1 (Ki-67) antigen. (e) shows details of the
       same spot containing stained and non-stained nuclei of normal
       as well as abnormal cells.}
     \label{fig:rcc}
  \end{center}
\end{figure*}

The tissue microarray (TMA) technology significantly accelerated
studies seeking for associations between molecular changes and
clinical endpoints \cite{Kononen98}. In this technology, $0.6 mm$
tissue cylinders are extracted from primary tumor material 
of hundreds of different patients and these
cylinders are subsequently embedded into a recipient tissue block.
Sections from such array blocks can then be used for simultaneous in
situ analysis of hundreds or thousands of primary tumors on DNA, RNA,
and protein level (cf. \ref{fig:rcc}). These results can then be
integrated with expression profile data which is expected to enhance
the diagnosis and prognosis of ccRCC \cite{Takahashi01, Moch99,
  Young01}. The high speed of arraying, the lack of a significant 
damage to donor blocks, and the regular arrangement of arrayed
specimens substantially facilitates automated analysis.
		      			
\begin{figure}[tb]
  \begin{center}	     	          	
    \includegraphics[width=0.7\linewidth] {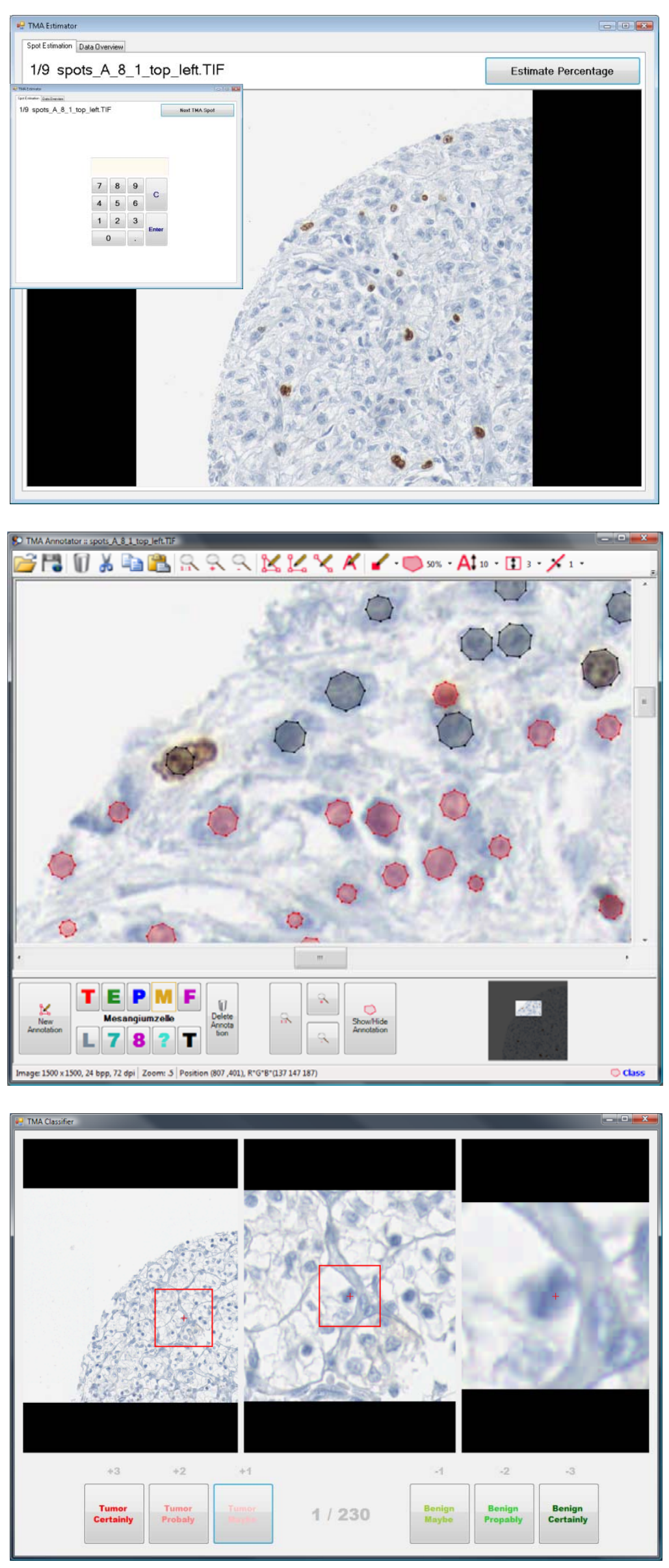}
    \caption{Tablet PC labeling applications for (i) global staining estimation
	(ii) nuclei detection and (iii) nuclei classification (from
      top to bottom).} 
    \label{fig:software}
  \end{center}
\end{figure}
			
Although the production of tissue microarrays is an almost routine
task for most laboratories, the evaluation of stained tissue
microarray slides remains tedious human annotation work, it is time
consuming and prone to error. Furthermore, the significant
intratumoral heterogeneity of RCC results in high interobserver
variability. The variable architecture of RCC also results in a
difficult assessment of prognostic parameters. Current image analysis
software requires extensive user interaction to properly identify cell
populations, to select regions of interest for scoring, to optimize
analysis parameters and to organize the resulting raw data. Because of
these drawbacks in current software, pathologists typically collect
tissue microarray data by manually assigning a composite staining
score for each spot - often during multiple microscopy sessions over a
period of days. Such manual scoring can result in serious
inconsistencies between data collected during different microscopy
sessions. Manual scoring also introduces a significant bottleneck that
hinders the use of tissue microarrays in high-throughput analysis.
		
\subsection{Analyzing Pathologists}\label{ssec:labelingex}
To assess the inter and intra variability of pathologists we designed
three different labeling experiments for the major tasks involved in
TMA analysis.  To facilitate the labeling process for trained
pathologists we developed software suite which allows the user to view
single TMA spots and which provides zooming and scrolling
capabilities.  The expert can annotate the image with vectorial data
in SVG (support vector graphics) format and he/she can mark cell nuclei,
vessels and other biological structures. In addition each structure
can be labeled with a class which is encoded  by its color.  To
increase usability and the adoption in hospitals we specifically
designed the software for tablet PC so that a pathologist can perform
all operations with a pen alone in a simple and efficient manner.
Figure \ref{fig:software} depicts the graphical user interfaces of the
three applications.

\begin{figure}[tb]
  \begin{center}	          	       
    \begin{tabular}{cc}     	                 
      \includegraphics[height=1.4in] {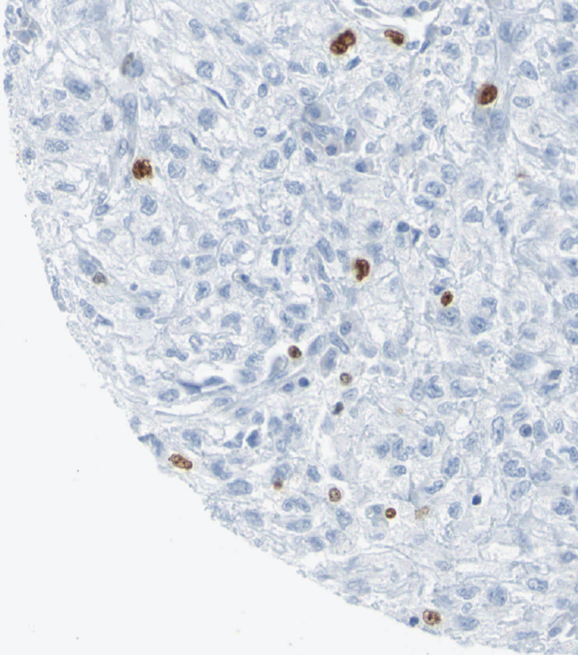} &
      \includegraphics[height=1.4in] {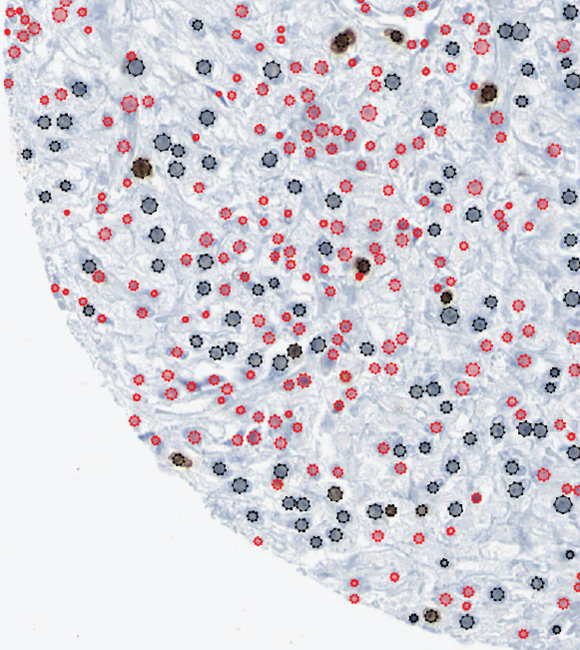} \\
      (a) & (b) \\
      \includegraphics[width=1.4in] {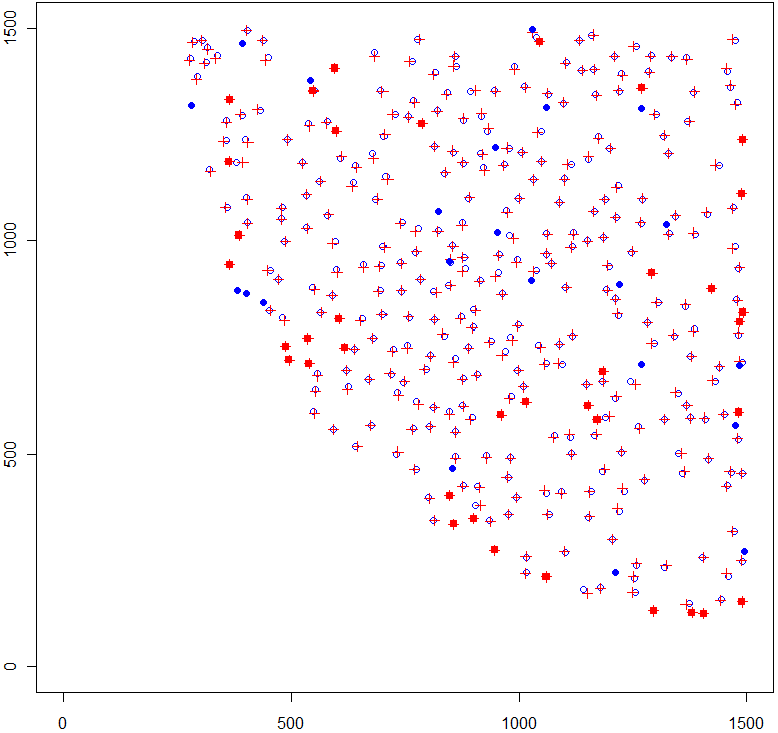} &
      \includegraphics[width=1.4in] {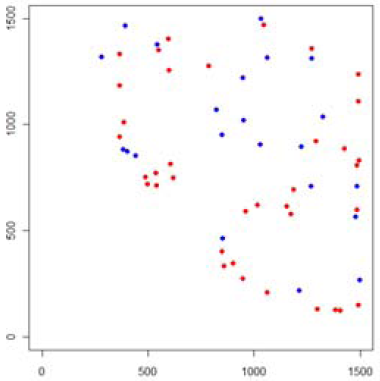} \\
      (c) & (d) \\
    \end{tabular}	
    \caption{(a) A quarter of an RCC TMA spot used for the nuclei detection experiment.
      (b) Annotations of one expert, indicating abnormal nuclei in black and normal ones in red.
      (c) Overlay of detected nuclei from expert one (blue circles)
      and expert two (red crosses). 
      (d) Disagreement between the two domain experts regarding the
      detection task. Nuclei which were labeled only by pathologist
      one are shown in blue and the nuclei found only by expert two
      are depicted in red.
    } 
    \label{fig:detectionvar}
  \end{center}
\end{figure}
				
\textbf{Nuclei Detection:} The most tedious labeling task is the
detection of cell nuclei.  In this experiment two experts on renal
cell carcinoma exhaustively labeled a quarter of each of the 9 spots
from the previous experiment.  Overall each expert independently
marked the center, the approximate radius and the class of more than
2000 nuclei.  Again a tablet PC was used so it was possible to split
up the work into several sessions and the experts could use the
machine at their convenience. 
The user detects nuclei by marking the location with the pen on the
tablet and indicates the diameter by moving the pen. 
A circular semi-transparent polygon is then drawn to mark the
nucleus. The final step consists of choosing a class for the nucleus.
In this setting it was either black for cancerous nuclei or red for
normal ones.  This task has to be repeated for each nucleus on each
spot. Finally it is possible to show and hide the annotation to gain
an overview over the original tissue.  Figure \ref{fig:detectionvar}
depicts a quarter of one of the RCC TMA spots together with the
annotation and the inter expert disagreement.
				 				
The average precision of one pathologist compared to the other is
$0.92$ and the average recall amounts to $0.91$.  These performance
numbers show that even detecting nuclei on an histological slide is by
far not an easy or undisputed task.

\textbf{Nuclei Classification:} The third experiment was designed to
evaluate the inter and intra pathologist variability for nuclei
classification, i.e. determining if a nucleus is normal/benign or
abnormal/malignant. This step crucially influences the final outcome due to
the fact that the percentage of staining is only estimated on the
subset of cancerous nuclei. In the experiment, $180$ randomly selected
nuclei are sequentially presented in three different views of varying
zoom stage. 
The query nucleus is indicated in each view with a red cross and the
area which comprises the next zoom view is marked with a red bounding
box (cf. Figure \ref{fig:software}).  During the setup phase the user
can adjust these views to simulate his usual workflow as good
as possible. During the experiment the expert has to select a class
for each nucleus and rate his confidence. Thus, he has the choice
between six buttons: tumor certainly, tumor probably, tumor maybe,
benign certainly, benign probably and benign maybe. After classifying
all nuclei, which have been classified as tumor, are displayed again
and the pathologist has to estimate if the nucleus is stained or
not. Again he has to rate his confidence in his own decision on a
scale of three levels.  To test the intra pathologist's variability a
subset of nuclei was queried twice but the images were flipped and
rotated by 90 degree at the second display to hamper recognition.
			
\begin{figure*}[tb]
  \begin{center}
    \begin{tabular}{cccc}
      \includegraphics[height=1.25in] {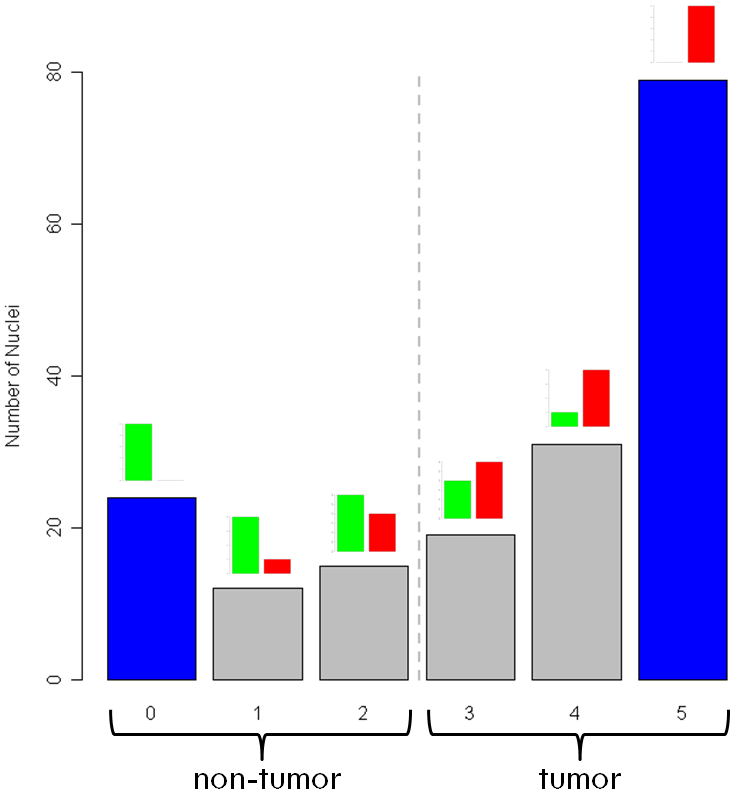} &
      \includegraphics[height=1.25in] {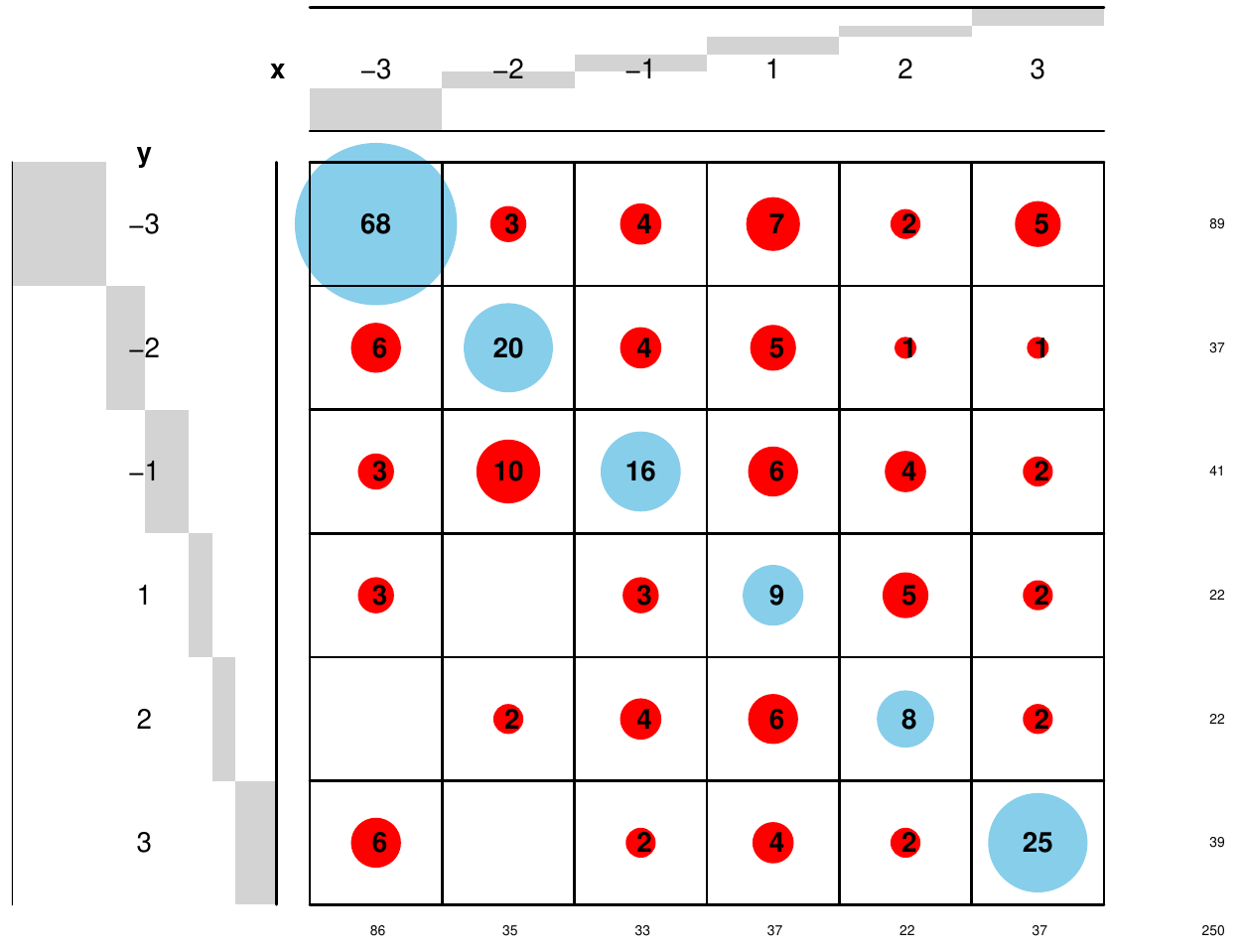} &
      \includegraphics[height=1.25in] {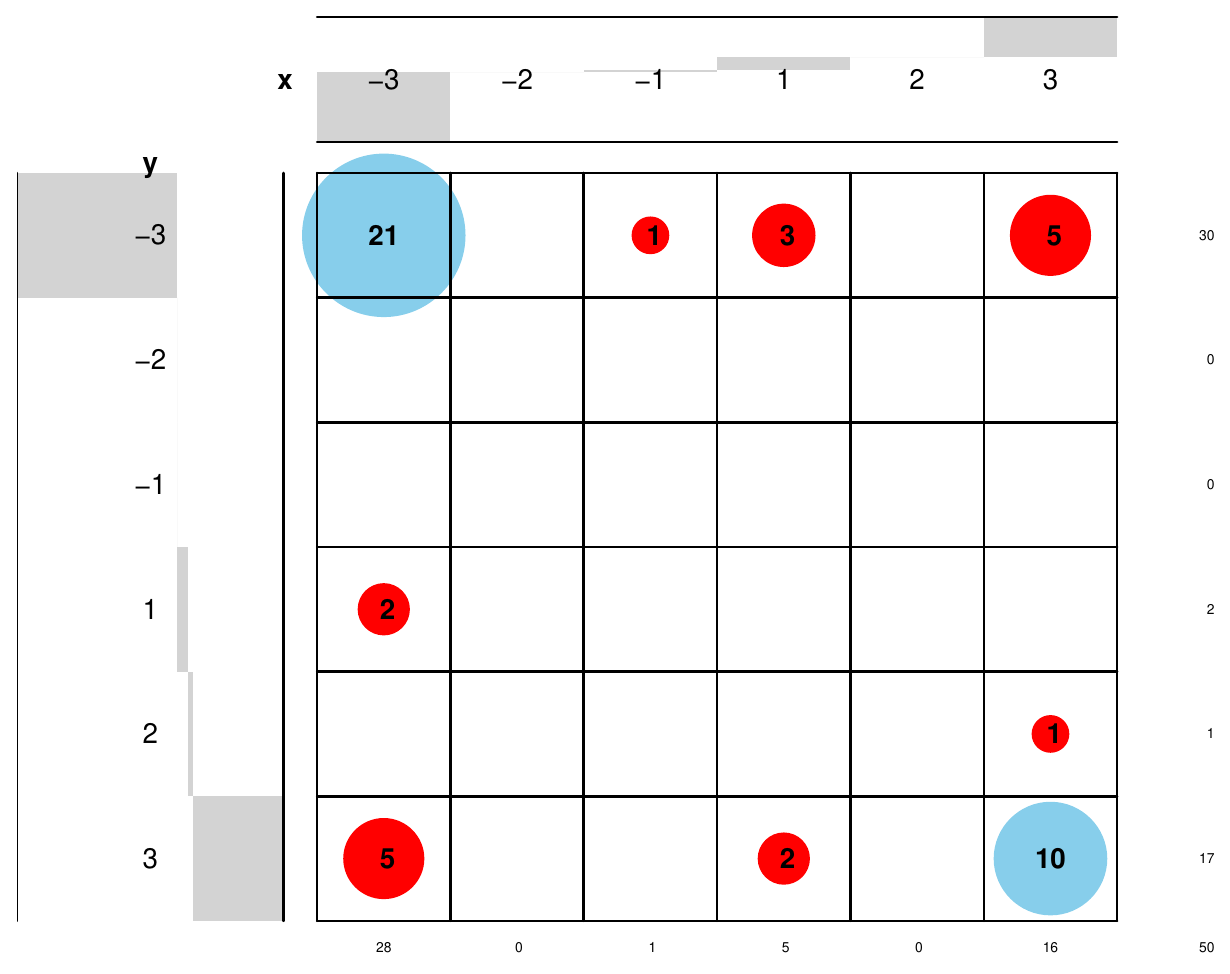} &
      \includegraphics[height=1.25in] {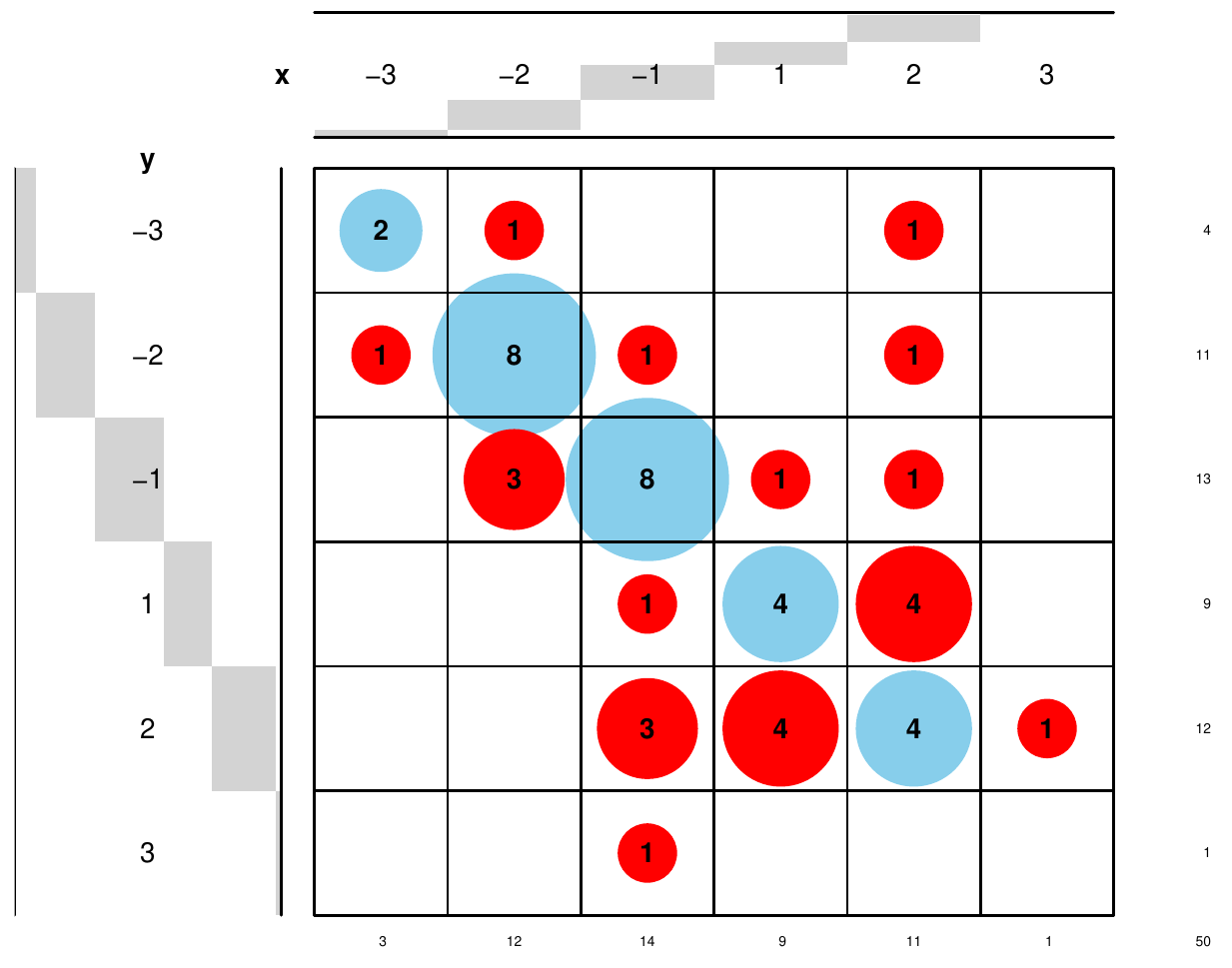} \\
      (a) & (b) & (c) & (d)\\
    \end{tabular}
    \caption{(a) Inter-pathologist classification variability based on
      $180$ nuclei labeled by five domain experts. The experts agree
      on $105$ out of $180$ nuclei (blue bars: $24$ normal, $81$
      cancerous).  (b-d) Confusion matrices including reader
      confidence for intra-observer variability in nuclei
      classification: (b) The combined result of all five experts
      yields a intra pathologist classification error of $21.2\%$.
      (c) Example of an extremely self-confident pathologist with
      $30\%$ error.  (d) A very cautions pathologist with a
      misclassification error of $18\%$.}
    \label{fig:baloonPathos}
  \end{center}
\end{figure*}
				
The results for inter-pathologist variability for the binary
classification task are plotted in Figure \ref{fig:baloonPathos}a. Out
of $180$ nuclei all five experts agreed on $24$ nuclei to be normal
and $81$ nuclei to be ab-normal, respectively cancerous. For the
other $75$ nuclei ($42\%$) the pathologists disagreed.
				
The analysis of the intra-pathologist error is shown in Figure
\ref{fig:baloonPathos}b. The overall intra classification error is
$21.2\%$ This means that every fifth nucleus was classified by an
expert first as cancerous and the second time as normal or vice versa.
The self-assessment of confidence allows us also to analyze single
pathologists.  For example Figure \ref{fig:baloonPathos}c shows the
results of a very self-confident pathologist who is always very
certain of his decisions but ends up with an error of $30\%$ in the
replication experiment. Figure \ref{fig:baloonPathos}d on the other
hand is the result of a very cautious expert who is rather unsure of
his decision, but with a misclassification error of $18\%$ he performs
significantly better than the previous one.  The important lesson
learned is, that self-assessment is not a reliable information 
to learn from. The intuitive notion, to use only training samples
which were classified with high confidence by domain experts is not
valid.
				
In defense of human pathologist it has to be mentioned that these
experiments represent the most general way to conduct a TMA analysis
and analogous studies in radiology report similar results
\cite{Saur2009,Saur2010}. In practice, domain experts focus only on
regions of TMA spots which are very well processed, which have no
staining artifacts or which are not blurred. The nuclei analyzed in
this experiment were randomly sampled from the whole set of detected
nuclei to mimic the same precondition which an algorithm would
encounter in routine work. Reducing the analysis to perfectly
processes regions would most probably decrease the intra-pathologist
error.

\textbf{Staining Estimation:} The most common task in manual TMA
analysis requires to estimate the staining. To this end a domain expert
views the spot of a patients for several seconds and estimates the
number of stained abnormal cells without resorting to actual nuclei
counting. This procedure is iterated for each spot on a TMA-slide to
get an estimate for each patient in the study.
				
It is important to note that, due to the lack of competitive
algorithms, the results of nearly all TMA studies are based on this
kind of subjective estimations.
				
\begin{figure*}[htb]
  \begin{center}	          	       
    \begin{tabular}{cc}     	                 
      \includegraphics[height=2.5in] {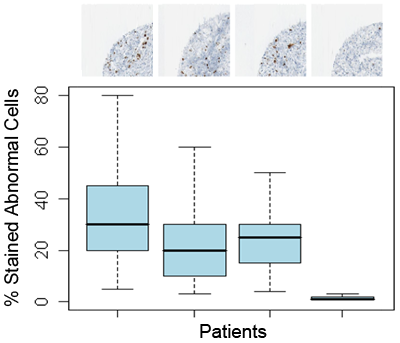} &
      \includegraphics[height=2in] {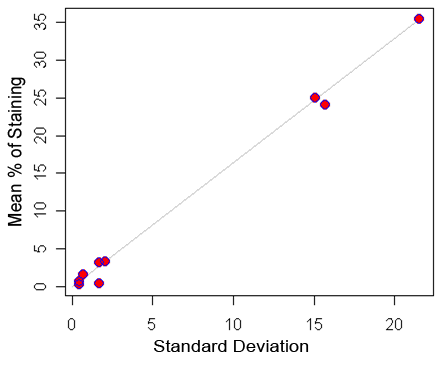} \\
      (a) & (b) \\
    \end{tabular}	
    \caption{(a) Results for 4 TMA spots from the labeling experiment
      conducted to investigate the inter pathologist variability for
      estimating nuclear staining. 14 trained pathologists estimated
      MIB-1 staining on 9 TMA spots.  The boxplots show a large
      disagreement between pathologist on spots with an averages
      staining of more than $10\%$.  The absolute estimated percentage
      is plotted on the y-axis. Spot 1 for example, yields a standard
      deviation of more than $20\%$. (b) The standard
      deviation grows linearly with the average estimated staining.  
    }
    \label{fig:stainingvar}
  \end{center}
\end{figure*}
					
To investigate estimation consistency we presented 9 randomly selected
TMA spots to 14 trained pathologists of the University Hospital
Zurich.

The estimations of the experts varied by up to $20\%$ as shown in
Figure \ref{fig:stainingvar}b.  As depicted in Figure
\ref{fig:stainingvar}b the standard deviation between the experts
grows linearly with the average estimated amount of staining.  The
high variability demonstrates the subjectivity of the estimation
process. This uncertainty is especially critical for types of cancer
for which the clinician chooses the therapy based on the estimated
staining percentage.  This result not only motivates but emphasizes
the need for more objective estimation procedures than current
practice. Our research is stimulated by the hope, that computational
pathology approaches do not only automated such estimation processes
but also produce better reproduceable and more objective results than
human judgment.

\subsection{Expert Variability in Fluorescence Microscopy}
Complementary to immunohistochemical TMA analysis, fluorescence
microscopy is applied 
often for high-throughput screening of molecular phenotypes. A
comprehensive study evaluating the performance of domain experts
regarding the detection of lymphocytes is presented by
\cite{Nattkemper03}.  In a best case, a medium-skilled expert needs on
average one hour for analyzing a fluorescence micrograph.  Each
micrograph contains between 100 and 400 cells and is of size $658
\times 517$ pixel.  Four exemplary micrographs were blindly evaluated
by five experts.  To evaluate the inter-observer variability
nattkemper et al. 
\cite{Nattkemper03} define a gold standard comprising all cell
positions in a micrograph that were detected by at least two experts.
		
Averaged over of CD3, CD4, CD7 and CD8 the sensitivity of the four
biomedical experts is varying between $67.5\%$ and $91.2\%$ and the
positive predictive value (PPV) between $75\%$ and $100\%$. Thus the
average detection error over all biomedical experts and micrographs is
approximately $17\%$.  Although fluorescence images appear to be
easier to analyze due to their homogeneous background, this high
detection error indicates the difficulty of this analysis task. These
results corroborates the findings in the ccRCC detection experiment
described in Section \ref{subsec:introRCC}.

\subsection{Generating a Gold Standard}\label{ssec:goldstandard}
The main benefit of labeling experiments like the ones described
before, is not to point out the high inter and intra variability
between pathologists, but to generate a gold standard.  In absence of
an objective ground truth measurement process, a gold standard is
crucial for the use of statistical learning,  first for learning a
classifier or regressor and second for validating the statistical
model.

Section \ref{sec:view} shows an example how the information gathered
in the experiments of Section \ref{ssec:labelingex} can be used to
train a computational pathology system.

Besides labeling application which are developed for specific
scenarios as the one described in Section \ref{ssec:labelingex}
several other possibilities exist to acquire data in pathology in a
structured manner.  Although software for tablet PCs is the most
convenient approach to gather information directly in the hospital it
is limited by the low number of test subjects which can complete an
experiment. To overcome this limitation the number of labelers can be
extended significantly by the use of web-based technologies.

Crowd-sourcing services like Amazon Mechanical Turk can be used to
gather large numbers of labels at a low cost.  Applications in
pathology suffer from the main problem, that the labelers are all
non-experts.  While crowd-sourcing works well for task based on
natural images \cite{Welinder10}, it poses a considerable problems in
pathology where for example the decision if a nucleus is normal or
cancerous is based on complicated criteria \cite{WHO04} which require
medical training and knowledge. Likewise the recognition of some
supercellular morphological structures requires years of training and
supervision.  Nevertheless crowd-sourcing could be useful in simple
detection tasks like finding nuclei in histological slides.

\subsection{Multiple Expert Learning}							
In classical supervised learning, a set of training data
$\{(x_i,y_i)\}_{i=1,\ldots,n}$ is available which consists of objects
$x_i$ and their corresponding labels $y_i$. The task is to predict the
label $y$ for a new test object $x$.  This approach is valid as long
as the target variable $Y=\{y_1,\ldots,y_n\}$ denotes
the ground truth of the application. If this condition is met, $Y$ and
$X=\{x_1,\ldots,x_n\}$ can be used for classifier learning and
evaluation.

Unfortunately, for a large number of real word application ground
truth is either not available or very expensive to acquire. In
practice, as a last resort, one would ask several domain experts for
their opinion about each object $x_i$ in question to generate a gold
standard as described in Section \ref{ssec:goldstandard}. Depending on
the difficulty of the task and the experience of the experts this
questioning often results in an ambiguous labeling due to disagreement
between experts. In pathology, very challenging scenarios, like
assessing the malignancy of a cell, not only the inter, but also the
intra expert variability is quite large (cf. Section
\ref{ssec:labelingex}).  Moreover, restricting the dataset to the
subset of consistently labeled samples results in loss of the majority
of data in these scenarios.

Consequently, 
such a data acquisition procedure poses a fundamental problem for
supervised learning.  Especially in computational pathology there is
clearly a need for novel algorithms to address the labeling problem and
to provide methods to validate models under such circumstances.

\begin{figure}[tb]
  \begin{center}
    \begin{tabular}{cc}
      (a) & \includegraphics[width=2.75in] {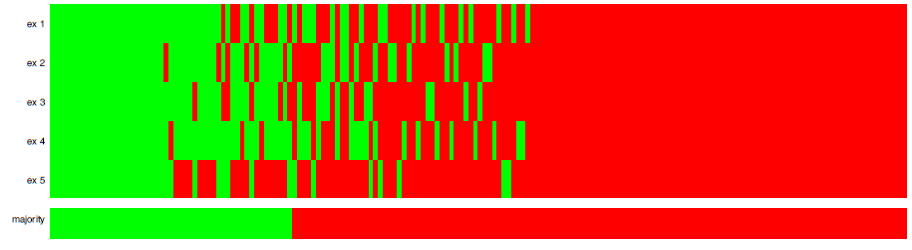} \\
      (b) & \includegraphics[width=2.75in] {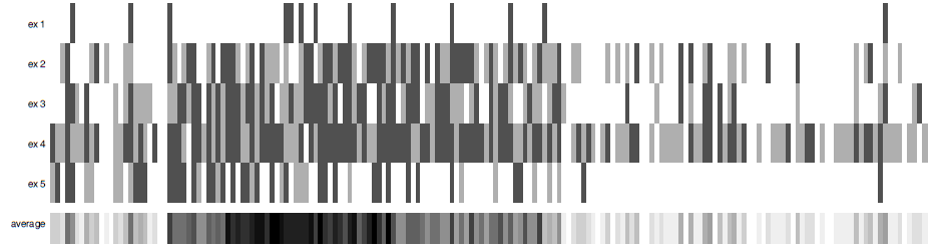} \\
    \end{tabular}
    \caption{Labeling matrix with majority vote (a) and confidence matrix with 
    	confidence average (b) of five domain experts classifying $180$ ccRCC nuclei
    	into cancerous (red) and benign (green).} 
    \label{fig:labelingmatrices}
  \end{center}
\end{figure}
		
More formally, each $y_i$ is replaced by a $D$ dimensional vector
$\bar{y_i}=\{y_i^1,\ldots,y_i^D\}$, where $y_i^d$ represents the
$i$-th label of domain expert $d$. To this end one is interested in
learning a classifier $\Phi(X,\bar{Y})$ from the design matrix $X$ and
the labeling matrix $\bar{Y}$.  To date it is an open research
question, how such classifier $\Phi(X,\bar{Y})$ should be formulated.
		
Recently \cite{Smyth94, Raykar09} presented promising results based on
expectation maximization where the hidden ground truth is estimated in
turn with the confidence in the experts.  Also along this lines
\cite{Whitehill09} introduced a probabilistic model to simultaneously
infer the label of images, the expertise of each labeler, and the
difficulty of each image. An application for diabetes especially for
detecting hard exudates in eye fundus images was published by
\cite{Kauppi09}.
					
Although a number of theoretical results exist \cite{Lugosi92,
  Smyth96, Dekel09}, empirical evidence is still lacking to establish
that these approaches are able to improve over simple majority voting
\cite{Tullock59, Downs61} in real world applications.

A further, promising line of research investigates the question if
such a classifier $\Phi(X,\bar{Y})$ can be learned in an on-line
fashion, especially when the new labels come from a different domain
expert. An affirmative answer
would show a high impact in domains where specific models can be
trained for years by a large number of experts, e.g. medical decision
support.
					
In summary, extending supervised learning to handle domain expert
variability is an exciting challenge and promises direct impact on
applications not only in pathology but in a variety of computer vision
tasks where generating a gold standard poses a highly non trivial
challenge.

\subsection{Public Datasets with Labeling Information}
		
The available of public datasets with labeling information is crucial
for the advance of an empirical science.  Although a comprehensive
archive like the UCI machine learning repository \cite{Frank10} does
not exist for computational pathology, there are a number of datasets
and initiatives which disseminate various kinds of data.

\textbf{Immunohistochemistry:} The most comprehensive database for
antibodies and human tissue is by far the Human Protein Atlas
\cite{Berglund08, Ponten08}. Comprising spots from tissue micro arrays
of $45$ normal human tissue types, it contains anywhere from $0-6$
images for each protein in each tissue type. The images are roughly 
$3000 \times 3000$ pixels in size, with each pixel approximately
representing a $0.5 \times 0.5 \mu m$ region on the microscopy slide.

A segmentation benchmark for various tissue types in bioimaging was
compiled by \cite{Manjunath09}, including 58 histopathological H\&E
stained images of breast cancer. The dataset provides labels from a
single expert for the tasks of segmentation and cell counting.

\textbf{Cytology:} Automation in cytology is the oldest and most
advanced branch in the field of image processing in pathology. Reason
therefore are that digital imaging is rather straightforward and that
single cells on a homogeneous background are more easily detected and
segmented than in tissue.  As a result commercial solutions are
available since decades.  Nevertheless especially the classification
of detected and segmented nuclei still poses large difficulties for
computational approaches.  \cite{Lezoray02} published ten color
microscopic images from serous cytology with hand segmentation
labels.\\ For bronchial cytology \cite{Meurie05} provide eight color
microscopic images.  Ground truth information for three classes
(nucleus, cytoplasm, and background pixels) is also available for each
image.  Pixels have a label specifying their classes (2: nucleus, 1:
cytoplasm, 0: background).\\ A dataset of $3900$ cells has been
extracted from microscopical image (serous cytology) by
\cite{Lezoray03}. This database has been classified into $18$ cellular
categories by experts.

\textbf{Fluorescence Microscopy:} A hand-segmented set of 97
fluorescence microscopy images with a total of 4009 cells has been
published by \cite{Murphy09}.  For fluorescence microscopy, the
simulation of cell population images is an interesting addition to
validation with manual labels of domain experts.  \cite{Nattkemper03,
  Lehmussola07} present simulation frameworks for synthetically
generated cell population images. The advantage of these techniques is
the possibility to control parameters like cell density, illumination
and the probability of cells clustering together.  \cite{Lehmussola07}
supports also the simulation of various cell textures and different
error sources.  The obvious disadvantage are (i) that the model can
only simulate what it knows and therefore can not represent the whole
variability of biological cell images and (ii) that these methods can
only simulate cell cultures without morphological structure. The later
disadvantage also prevents their use in tissue analysis.  Although the
thought of simulated tissue images in light microscopy is appealing,
currently there does not exist methods which could even remotely
achieve that goal.

\section{Imaging: From Classical Image Processing to Statistical Pattern Recognition}\label{sec:imaging}

In recent years, a shift from rule based expert system towards learned
statistical models could be observed in medical information
systems. The substantial influence that machine learning had on the computer
vision community is also reflecting more and more on medical imaging
in general and histopathology in special. Classifiers for object
detection and texture description in conjunction with various kinds of
Markov random fields are continuously replacing traditional watershed
based segmentation approaches and handcrafted rule-sets.
Just recently \cite{Monaco10} successfully demonstrated the use of
pairwise Markov models for high-throughput detection of prostate cancer 
in histological sections.
An excellent review of state-of-the-art histopathological image analysis 
methodology was compiled by \cite{Gurcan09}.
		
As with most cutting edge technologies, commercial imaging solutions
lag behind in development but the same trend is evident. \cite{Rojo09}
review commercial solutions for quantitative immunohistochemistry in
the pathology daily practice.
	
Despite the general trend towards probabilistic models, very classical
approaches like mathematical morphology \cite{Soille03} are still used
with great success. Recently, \cite{Lezoray09} presented a framework
for segmentation based on morphological clustering of bivariate color
histograms and \cite{FuchsDAGM2008} devised an iterative morphological
algorithm for nuclei segmentation.
		
Besides common computer vision tasks like object detection,
segmentation and recognition, histopathological imaging poses domain
specific problems such as estimating staining of nuclei conglomerates
\cite{Grabe09} and differentiation nuclei by their shape \cite{Arif07}.

\subsection{Preprocessing vs. Algorithmic Invariance}\label{ssec:invariance}

Brightfield microscopic imaging often produces large differences of
illumination within single slides or TMA spots. These variations are
caused by the varying thickness of the slide or by imperfect
staining. Such problems can be overcome either by preprocessing the
image data or by designing and integrating invariance into the
algorithmic processing to compensate for these variations.

Inconsistencies in the preparation of histology slides render it
difficult to perform a quantitative analysis on their results.
An normalization approach based on optical density and SVD projection
is proposed by \cite{Macenko09} for overcoming some of the
known inconsistencies in the staining process. Slides which were 
processed or stored under very different conditions are projected 
into a common, normalized space to enable improved quantitative analysis.

Preprocessing and normalization methods usually not only reduce noise
induced differences between samples but often also eliminate the
biological signal of interest. As an alternative to such an
irreparable information loss during data acquisition, algorithms with
illumination invariance or with compensation of staining artifacts are
designed which are robust to these uncontrollable experimental
variations. 
		
Relational Detection Forests \cite{FuchsISVC2009} provide one possibility
to overcome this problem of information loss. 
Especially designed for detection of cell nuclei
in histological slides, they are based on the concept of randomized
trees \cite{Breiman01}. The features, which are selected for this
framework center around the idea that relation between features are
more robust than thresholds on single features. A similar idea was
applied by \cite{Geman04} to gene chip analysis where similar problems
occur, due to the background noise of different labs.  Contrary to the
absolute values the relation between DNA expression is rather robust.

Object detection is commonly solved by training a classifier on
patches centered at the objects of interest \cite{Viola01}, e.g., the
cell nuclei in medical image processing of histological slides.
Considering only the relation between rectangles within these patches
results in illumination invariant features which give the same
response for high and low contrast patches as long as the shape of the
object is preserved. It has to be noted, that due to the
directionality of the relation they fail if the image is inverted. In
general, illumination invariance speeds up the whole analysis process
because neither image normalization nor histogram equalization are
required.
		
The feature base proposed in \cite{FuchsISVC2009} is defined as
follows:  The coordinates of two rectangles $R_1$ and $R_2$ are
sampled uniformly within a predefined window size $w$:
\[ 
R_i=\{c_{x1},c_{y1},c_{x2},c_{y2}\},\quad c_i \sim  U(x|0,w)
\]
For each rectangle the intensities of the underlying gray scale image
are summed up and normalized by the area of the rectangle.  The
feature $f(s,R_1,R_2)$ evaluates to a boolean value by comparing these
quantities:
\begin{equation}
f(s,R_1,R_2) = \begin{cases} 1 & \text{if }
  \displaystyle\sum_{i|x_i \in  R_1}  \frac{x_i}{n_{1}} < 
  \sum_{i|x_i \in R_2}\frac{x_i}{n_{2}} \\
  0 & \text{otherwise}
\end{cases}
\end{equation}
where $x_i$ is the gray value intensity of pixel $i$ of sample
$s=\{x_1,x_2,\ldots,x_n\}$ and 
$n_1,n_2$ denote the number of samples in $R_1, R_2$, respectively.
From a general point of view this definition is similar to generalized
Haar features but there are two main differences: (i) the quantity of
interest is not the continuous difference between the rectangles but
the boolean relation between them and hence (ii) it is not necessary
to learn a threshold on the difference to binarize the feature.
						
For example, in the validation experiments a window size of
$65\times65$ pixels was chosen.  Taking into account that rectangles
are flipping invariant, this results in
$\left((64^4)/4\right)^2 \approx 2\cdot10^{13}$ possible
features.\\ Putting this into perspective, the restriction of the
detector to windows of size $24\times 24$ leads to $\sim6.9\cdot10^9$
features which are significantly more than the $45,396$ Haar features
from classical object detection approaches \cite{Viola01}.
		
For such huge feature spaces it is currently not possible to
exhaustively evaluate all features while training a
classifier. Approaches like AdaBoost \cite{freund96} which yield very
good results for up to hundreds of thousands of features are not
applicable any more.  These problems can be overcome by employing
randomized algorithms \cite{FuchsISVC2009, Geurts06} where features
are sampled randomly for learning classifiers on these random
projections.

\subsection{Inter-Active and Online Learning for Clinical
  Application}\label{ssec:interactive} 

Day-to-day clinical application of computational pathology algorithms
require adaptivity to a large variety of scenarios. 
Not only that staining protocols and slide
scanners are constantly updated and changed but common algorithms like
the quantification of proliferation factors have to work robustly on
various tissue types.  The detection of multiple objects like nuclei
in noisy images without an explicit model is still one of the most
challenging tasks in computer vision.  Methods which can be applied in
an plug-and-play 
manner are not available to date.

\cite{FuchsOLCV2009} present an inter-active ensemble learning
algorithm based on randomized trees, which can be employed to learn an
object detector in an inter-active fashion. In addition this learning
method can cope with high dimensional feature spaces in an efficient
manner and in contrast to classical approaches, subspaces are not
split based on thresholds but by learning relations between features.

Incorporating the knowledge of domain experts into the process of
learning statistical models poses one of the main challenges in machine
learning \cite{Vapnik98} and computer vision.  Data analysis applications in
pathology share properties 
of online and active learning which can be termed inter-active
learning. The domain expert interferes with the learning process by
correcting falsely classified samples.  Algorithm
\ref{alg_interactive} sketches an overview of the inter-active
learning process.
		 
In recent years online learning has been of major interest to a large
variety of scientific fields.  From the viewpoint of machine learning
\cite{Blum96} summarizes a comprehensive overview of existing methods and
open challenges. In computer vision online boosting has been
successfully applied to car detection \cite{Nguyen07}, video
surveillance \cite{celik08} and visual tracking \cite{Grabner08}.  One
of the first inter-active frameworks was developed by \cite{roth08}
and applied to pedestrian detection.
		
Ensemble methods like boosting \cite{freund96} and random forests
\cite{Amit97, Breiman01} celebrated success
in a large variety of tasks in statistical learning but in most cases
they are only applied offline. Lately, online ensemble learning for
boosting and bagging was investigated by \cite{oza01} and
\cite{Fern03}.  The online random forest as proposed by
\cite{Elgawi08} incrementally adopts new features. Updating decision
trees with new samples was described by \cite{utgoff89,utgoff94} and
extended by \cite{kalles96,PfahringerHK07}. Update schemes for pools
of experts like the WINNOW and Weighted Majority Algorithm were
introduced by \cite{Littlestone88,Littlestone89} and successfully
employed since then.
		
In many, not only medical domains, accurate and robust object
detection specifies a crucial step in data analysis pipelines. In
pathology for example, the detection of cell nuclei on histological
slides serves as the basis for a larger number of tasks such as
immunohistochemical staining estimation and morphological grading.
Results of medical interest such as survival prediction are
sensitively influenced by the accuracy of the object detection
algorithm. The diagnosis of the pathologist in turn leads to different
treatment strategies and hence directly affects the patient.  For most
of these medical procedures the ground truth is not known (see Section
\ref{sec:ground.truth}) and for most problems biomedical science lacks
orthogonal methods which could verify a considered
hypotheses. Therefore, the subjective opinion of the medical doctor is
the only gold standard available for training such decision support
systems.

\begin{algorithm}[htbp]
  \SetAlgoInsideSkip{medskip}
  \vspace{0.3cm}
  \KwData{Unlabeled Instances $U = \{u_1,\ldots,u_n\}$ }	 \%(e.g. image)
  \KwIn{Domain Expert $E$}
  \KwOut{Ensemble Classifier $C$}		
  \caption{Schematic workflow of an inter-active ensemble learning framework. 
    The domain expert interacts with the algorithm
    to produce a classifier (object detector) which satisfies the conditions 
    based on the experts domain knowledge. }
  
  \vspace{0.3cm}
  \While{(expert is unsatisfied with current result)}
	{
	  classify all samples $u_i$\;
	  \While{(expert corrects falsely predicted sample $u_i$ with label $l_i$)}
		{					
		  update weights of the base classifiers\\
		  learn new base classifiers
		}						
	}
	return $C$
	
	\label{alg_interactive}
\end{algorithm}

In such scenarios the subjective influence of a single human can be
mitigated by combining the opinions of a larger number of experts. In
practice consolidating expert judgments is a cumbersome and expensive
process and often additional experts are not available at a given
time. To overcome these problems online learning algorithms are
capable of incorporating additional knowledge, so-called
side-information, when it is available.
		
In an ideal clinical setting, a specialized algorithm for cell nuclei
detection should be available for each subtype of cancer. By using and
correcting the algorithm several domain experts as its users
continuously train and update the method. Thereby, the combined knowledge of a
large number of experts and repeated training over a longer period of
time yields more accurate and more robust classifiers than batch
learning techniques.
		
The described setting differs from the conventional views of online
learning and active learning insofar that new samples are neither
chosen at random nor proposed for labeling by the algorithm itself. In
addition, the adversary is not considered malicious but also not
completely trustworthy. The domain expert reacts to the classification
of unlabeled data and corrects wrongly classified instances.  These
preconditions lead to the success or failure of different combination rules.
		
It has to be noted, that these kind of machine learning approaches are
in sharp contrast to classical rule bases expert systems
\cite{Hayes83} which are still used by a number of commercial medical
imaging companies.  For these applications the user has to be an image
processing experts who chooses dozens of features and thresholds by
hand to create a rule set adapted to the data.  Contrary to that
strategy, in an inter-active learning framework the user has to be a
domain expert, in our case a trained pathologists. Feature extraction
and learning of statistical models is performed by the algorithms so
that the expert can concentrate on the biomedical problem at hand.
Inter-active learning frameworks like \cite{Nguyen07, FuchsOLCV2009}
show promising results, but further research especially on long term
learning and robustness is mandatory to estimate the reliability of
these methods prior to an application in clinical practice.

\subsection{Multispectral Imaging and Source Separation}
		
Multispectral imaging \cite{Levenson2006, Loos08} for
immunohistochemically stained tissue and brightfield microscopy seems
to be a promising technology although a number of limitations have to
be kept in mind.
			
To date, double- or triple-staining of tissue samples on a single
slide in brightfield (non-fluorescence) microscopy poses still a major
challenge. 
Traditionally, double staining relied on chromogens, which have been
selected to provide maximum color contrast for observation with the
unaided eye.  For visually good color combinations, however,
technically feasible choices always include at least one diffuse
chromogen, due to the lack of appropriate chromogen colors.
Additional problems arise from spatial overlapping and from unclear
mixing of colors.  Currently, these problem are addressed by cutting
serial sections and by staining each one with a different antibody and
a single colored label. Unfortunately, localized information on a
cell-by-cell basis is lost with this approach. In the absence of
larger structures like glands, registration of sequential slices
proved to be highly unreliable and often not feasible at all.
Multispectral imaging yield single-cell-level multiplexed imaging of
standard Immunohistochemistry in the same cellular compartment. This
technique even works in the presence of a counterstain and each label
can be unmixed into separate channels without bleed-through.
						
Computational pathology algorithms would profit from multispectral
imaging also is experiments with single stains, due to the possibility
to accurately separate the specific label signals from the background
counterstain.
			
Practical suggestions for immunoenzyme double staining procedures for
frequently encountered antibody combinations like rabbit–mouse,
goat–mouse, mouse–mouse, and rabbit–rabbit are discussed in
\cite{Loos08}. The suggested protocols are all suitable for a
classical red-brown color combination plus blue nuclear
counterstain. Although the red and brown chromogens do not contrast
very well visually, they both show a crisp localization and can be
unmixed by spectral imaging.

Detection an segmentation of nuclei, glands or other structures
constitute a crucial steps in various computational pathology
frameworks. With the use of supervised machine learning techniques
these tasks are often performed by trained classifiers which assign
labels to single pixels. Naturally one can ask if MSI could improve
this classification process and if the additional spectral bands
contain additional information?  A study conducted by
\cite{Boucheron07} set out to answer this question in the scope of
routine clinical histopathology imagery. They compared MSI stacks with
RGB imagery with the use of several classifier ranging from linear
discriminant analysis (LDA) to support vector machines (SVM).  For
H\&E slide the results indicate performance differences of less than
1\% using multispectral imagery as opposed to preprocessed RGB
imagery.  Using only single image bands for classification showed that
the single best multispectral band (in the red portion of the
spectrum) resulted in a performance increase of $0.57\%$, compared to
the performance of the single best RGB band (red).  Principal
components analysis (PCA) of the multispectral imagery indicated only
two significant image bands, which is not surprising given the
presence of two stains.  The results of \cite{Boucheron07} indicate
that MSI provides minimal additional spectral information than would
standard RGB imagery for routine H\&E stained histopathology.
			
Although the results of this study are convincing it has to be noted
that only slides with two channels were analyzed. For triple and
quadruple staining as described in \cite{Loos08} MSI could still
encode additional information which should lead to a higher
classification performance.  Similar conclusions are drawn by
\cite{Cukierski09}, stating that MSI has significant potential to
improve segmentation and classification accuracy either by
incorporation of features computed across multiple wavelengths or by
the addition of spectral unmixing algorithms.  
			
Complementary to supervised learning as described before,
\cite{Rabinovich03} proposed unsupervised blind source separation for
extracting the contributions of various histological stains to the
overall spectral composition throughout a tissue sample. As a
preprocessing step all images of the multispectral stack were
registered to each other considering affine transformations.
Subsequently it was shown that Non-negative Matrix Factorization (NMF)
\cite{Lee99} and Independent Component Analysis (ICA)
\cite{Hyvarinen01} compare favorable to Color Deconvolution
\cite{Ruifrok01}.  Along the same lines \cite{Begelman09} advocate
principal component analysis (PCA) and blind source separation (BSS)
to decompose hyperspectral images into spectrally homogeneous
compounds.

In the domain of fluorescence imaging \cite{Zimmermann05} give an
overview of several source separation methods. The main difficulty
stems from the significant overlap of the emission spectra even with
the use of fluorescent dyes.  To this end \cite{Newberg09} conduct a
study on more than 3500 images from the Human Protein Atlas
\cite{Berglund08, Ponten08}.  They concluded that subcellular
locations can be determined with an accuracy of $87.5\%$ by the use of
support vector machines and random forests \cite{Amit97, Breiman01}.
Due to the spread of Type-2 diabetes there is growing interest in
pancreatic islet segmentation and cell counting of $\alpha$ and
$\beta$-cells \cite{Herold09}.  An approach which is based on the
strategies described in Section \ref{ssec:invariance} and Section
\ref{ssec:interactive} is described in \cite{FlorosMICCAI09}.

It is an appealing idea to apply source separation techniques not only
to multispectral imaging but also to standard RGB images. This
approach could be useful for a global staining estimation of the
separate channels or as a preprocessing step for training a
classifier. Unfortunately, antigen-antibody reactions are not
stoichiometric. Hence the intensity/darkness of a stain does not
necessarily
correlate with the amount of reaction products. With the
exception of Feulgen staining also most histological stains 
are not stoichiometric.  \cite{Loos08} also state that the brown DAB
reaction product is not a true absorber of light, but a scatterer of
light, and has a very broad, featureless spectrum. This optical
behavior implies that DAB does not follow the Beer-Lambert law, which
describes the linear relationship between the concentration of a
compound and its absorbance, or optical density. As a consequence,
darkly stained DAB has a different spectral shape than lightly stained
DAB. Therefore attempting to quantify DAB intensity using source
separation techniques is not advisable.  Contrary to this observation,
employing a non-linear convolution algorithm as preprocessing for a linear
classifier, e.g. for segmentation could be of benefit.

Finally, multispectral imaging is not available for automated whole
slide scanning which constrains its applicability. Imaging a TMA
manually with a microscope and a MSI adapter is too tedious and time
consuming.

\subsection{Software Engineering Aspects}			
		
One of the earliest approaches for high performance computing in
pathology used image matching algorithms based on decision trees to
retrieve images from a database \cite{Wetzel97}. The approach was
applied to Gleason grading in prostate cancer. Web-based data
management frameworks for TMAs like \cite{Thallinger07} facilitate not
only storage of image data but also storage of experimental and
production parameters throughout the TMA workflow.

A crucial demand on software engineering is the ability to scale
automated analysis to multiple spots on a TMA slide and even
multiple whole microscopy slides.
Besides cloud computing one possibility to achieve that goal 
is grid computing.
\cite{Foran09} demonstrated the feasibility of such a system
by using the caGrid infrastructure \cite{Oster08} for Grid-enabled 
deployment of an automated cancer tissue segmentation algorithm for 
TMAs.

A comprehensive list of open source and public domain software for
image analysis in pathology is available at
\href{www.computational-pathology.org}{www.computational-pathology.org}.

\section{Statistics: Survival Analysis and Machine Learning in Medical Statistics}\label{sec:statistics}  
		
The main thrust of research in computational pathology is to build
completely probabilistic models of the complete processing pipelines
for histological and medical data. In
medical research this nearly always also includes time to event data,
where the event is either overall survival, specific survival, event
free survival or recurrence free survival of patients.  Statistics and
machine learning within this scope is defined as Survival Analysis.

\subsection{Censoring and Descriptive Statistics}		
Most difficulties in survival statistics arise from the fact, that
nearly all clinical datasets contain patients with censored survival
times. The most common form of censoring is right censored data which
means that the death of the patient is not observer during the runtime
of the study or that the patient withdrew from the study, e.g. because
he moved to another location.
		
The nonparametric Kaplan-Meier estimator \cite{Kaplan58} is frequently
used to estimate the survival function from right censored data. This
procedure requires first toorder the survival times from the smallest
to the largest such that $t_{1}\leq t_{2}\leq t_{3}\leq\ldots\leq
t_{n}$, where $t_{j}$ is the $j$th largest unique survival time. The
Kaplan-Meier estimate of the survival function is then obtained as
\begin{equation}
  \hat{S}(t) = \prod_{j:t_{(j)}\leq t}{\left(  1-\frac{d_j}{r_j}\right)}
\end{equation}
where $r_j$ is the number of individuals at risk just before $t_{j}$,
and $d_j$ is the number of individuals who die at time $t_{j}$.

To measure the goodness of separation between two or more groups, the
log-rank test (Mantel-Haenszel test) \cite{Mantel59} is employed to
assesses the null hypothesis that there is no difference in the
survival experience of the individuals in the different groups. The
test statistic of the log-rank test (LRT) is $\chi^2$ distributed:
\begin{equation}
  \hat{\chi^2} =
  \frac{
    \left(\sum^m_{i=1}{(d_{1i}-\hat{e}_{1i})}\right)^2
  }{
    \sum^m_{i=1}{\hat{v}_{1i}}
  } 
\end{equation}
where $d_{1i}$ is the number of deaths in the first group at $t_{i}$
and $e_{1i}=r_{1j}\frac{d_{i}}{r_i}$ where $d_i$ is the total number
of deaths at time $t_{(i)}$, $r_j$ is the total number of individuals
at risk at this time, and $r_{1i}$ the number of individuals at risk
in the first group. Figure \ref{fig:surv} depicts Kaplan-Meier plots
for two subgroups each and the LRT p-values. The associated data is
described in detail in Section \ref{sec:view}.

\subsection{Survival Analysis}
Survival Analysis as a branch of statistics is not restricted to
medicine but analyses time to failure or event data and is also
applicable to biology, engineering, economics etc.  Particularly in
the context of medical statistics, it is a powerful tool for
understanding the effect of patient features on survival patterns
within specific groups \cite{Klein97}. A parametric approach to such
an analysis involves the estimation of parameters of a probability
density function which models time.

In general the distribution of a random variable $T$ (representing
time) is defined over the interval $[0,\infty)$. Furthermore, a
standard survival function is specified based on the cumulative
distribution over $T$ as follows:

\begin{equation}
  S(t) = 1 - p(T \leq t_0) = 1 - \int_0^{t_0}{p(t)dt},
  \label{eq:survivalfunction}
\end{equation}
which models the probability of an individual surviving up to time
$t_0$.  The hazard function $h(t)$, the instantaneous rate of failure
at time $t$, is defined as follows:
\begin{equation}
  h(t) = \lim\limits_{\Delta t \rightarrow 0} \frac{P(t< T \leq t +
    \Delta t | T > t )}{\Delta t } = \frac{p(T=t)}{S(t)}. 
  \label{eq:hazardrate}
\end{equation}
		
The model is further extended by considering the effect of covariates
$X$ on time via a regression component.  In medical statistics the
most popular method for modeling such effects is Cox's proportionality
hazards model \cite{Cox72}:
\begin{equation}
  h(t | \bs x) = h_0 (t) \exp(\bs x^{T} \bs \beta ),
  \label{eq:hazards}
\end{equation}
where $h_0 (t)$ is the baseline hazard function, which is the chance
of instant death given survival till time $t$, $\bs x$ is the vector
of covariates and $\bs \beta$ are the regression coefficients.

\subsection{A Bayesian View of Survival Regression}			
		
Bayesian methods are gaining more and more popularity in machine
learning in general and in medical statistics in special. A big
advantage in survival analysis is the possibility to investigate the
posterior distribution of a model.  Especially in regularized survival
regression models \cite{RothFuchs08} it is possible to get a posterior
distribution also on zero coefficients, i.e. for biomarkers which
hence were not included in the model.
		
A common choice of distribution for modeling time is the Weibull
distribution which is flexible in terms of being able to model a
variety of survival functions and hazard rates. Apart from
flexibility, it is also the only distribution which captures both the
accelerated time model and the proportionality hazards model
\cite{josephming}.  The Weibull distribution is defined as follows:
\begin{equation}
  p(t | \alpha_w, \lambda_w)	 = \alpha_w \frac{1}{\lambda_w}
  t^{\alpha_w -1} \exp\left(-\frac{1}{\lambda_w} t^{\alpha_w}\right), 
  \label{eq:weib}
\end{equation}
where $\alpha_w$ and $\lambda_w$ are the shape and scale parameters,
respectively. Based on the above definition and assuming
right-censored data \cite{Klein97}, the likelihood assumes the form
\begin{equation}
  p(\left\{t_i\right\}_{i=0}^{N} | \alpha_w,\lambda_w) = 
  \prod_{i=1}^N{\left(\frac{\alpha_w}{\lambda_w} t_i^{\alpha_w
      -1}\right) ^{\delta_i} \exp\left(-\frac{1}{\lambda_w}
    t_i^{\alpha_w}\right)}, 
  \label{eq:like1}
\end{equation}
where $\delta_i = 0$ when the $i^{th}$ observation is censored and $1$
otherwise.  Further, to model the effect of covariates $\bs x$ on the
distribution over time, Cox's proportional hazards model can be
applied. Under this model, the covariates are assumed to have a
multiplicative effect on the hazard function.

\subsection{Higher Order Interactions}

A reoccurring question in biomedical research projects and especially
in TMA analysis studies interactions of markers and their influence on
the target. Two modern approaches within the scope of computational
pathology try to solve this question from a frequentist
\cite{Dahinden10} and a Bayesian \cite{RothFuchs08} point of view.
		
The most frequent approach for modeling higher-order interactions
(like pairs or triplets of features etc.) instead of modeling just the
main effects (individual features) are polynomial expansions of
features. For example the vector $\bs x = \left\{x_1,x_2,x_3\right\}$
can be expanded up to order 2 as $\bs x^{'} =
\left\{x_1,x_2,x_3,x_1:x_2,x_1:x_3,x_2:x_3,x_1:x_2:x_3\right\}$.
Additional flexibility is built into this model by including a random effect
in $\eta$ in the following manner:
\begin{equation}
  \eta = \bs x^\transp \bs \beta + \epsilon, \hspace{20pt} \text{where
    \  } \epsilon \sim N(0,\sigma^2). 
  \label{eq:etanormal}
\end{equation}
		
To include the covariate effect the likelihood of Equation
\ref{eq:like1} is modified as follows:
		
\begin{eqnarray*}
  p(\left\{t_i \right\}_{i=0}^{N} |\bs x_i, \alpha_w,\lambda_w) & = &
  \prod_{i=1}^N{\left[\frac{\alpha_w}{\lambda_w} t_i^{\alpha_w -1}
      \exp(\eta_i)\right] ^{\delta_i} } \cdot \\ 
  & & \dot
  \exp\left(-\frac{1}{\lambda_w} t_i^{\alpha_w} \exp(\eta_i)\right)
  \label{eq:like2}	
\end{eqnarray*}

These kind of models can be seen as enhancement of generalized linear
models \cite{Mccullaghand} and are called random-intercept models. For
a full Bayesian treatment of the model, suitable priors have to be
defined for the parameters of the model, namely $\alpha_w$,
$\lambda_w$, $\sigma$ and $\bs \beta$.  Useful priors for this model
are described in \cite{RothFuchs08}.

\subsection{Mixtures of Survival Experts}
		
Frequently,  sub-groups of patients specified by characteristic
survival times have to be identified together with the effects of
covariates within each sub-group. Such information might hint at the
disease mechanisms. Statistically this grouping is represented by
a mixture model or specifically by a mixture of survival experts.

To this end, \cite{rosentanner99} define a \textit{finite}
mixture-of-experts model by maximizing the partial likelihood for the
regression coefficients and by using some heuristics to resolve the
number of experts in the model. More recently
\cite{Ando04kernelmixture} use a maximum likelihood approach to infer
the parameters of the model and the Akaike information criterion (AIC)
to determine the number of mixture components.
		
A Bayesian version of the mixture model \cite{Kottas2006578} analyzes
the model with respect to time but does not capture the effect of
covariates. On the other hand the work by
\cite{Ibrahim96bayesianvariable} performs variable selection based on
the covariates but ignores the clustering aspect of the modeling.
Similarly, \cite{paserman04} defines an infinite mixture model but
does not include a mixture of experts, hence implicitly assuming that all
the covariates are generated by the same distribution with a common
shape parameter for the Weibull distribution.

\cite{RothFuchs08} unify the various important elements of this
analysis into a Bayesian mixture-of-experts (MOE) framework to model
survival time, while capturing the effect of covariates and also
dealing with an unknown number of mixing components. To infer the
number of experts a Dirichlet process prior on the mixing proportions
is applied, which solves the issue of determining the number of
mixture components beforehand \cite{Rasmussen02infinitemixtures}. Due
to the lack of fixed-length sufficient statistics, the Weibull
distribution is not part of the exponential family of distributions
and hence the regression component, introduced via the proportionality
hazards model, is non-standard. Furthermore, the framework of
\cite{RothFuchs08} includes sparsity constraints to the regression
coefficients in order to determine the key explanatory factors
(biomarkers) for each mixture component. Sparseness 
is achieved by utilizing a Bayesian version of the Group-Lasso
\cite{Raman09a, Raman09b} which is a sparse constraint for grouped
coefficients \cite{yuan06model}.

\section{The Computational Pathology Pipeline: A holistic View}\label{sec:view}
			
This chapter describes an genuine computational pathology project,
which has been designed following the principles described in the previous
sections. It is an ongoing project in kidney cancer research
conducted at the University Hospital Zurich and ETH Zurich. Parts of
it were published in \cite{FuchsMICCAI2008} and \cite{FuchsISVC2009},
where also algorithmic details of the computational approach can be
found.
			
\begin{figure*}[htbp]
  \begin{center}					
    \includegraphics[width=1\linewidth]{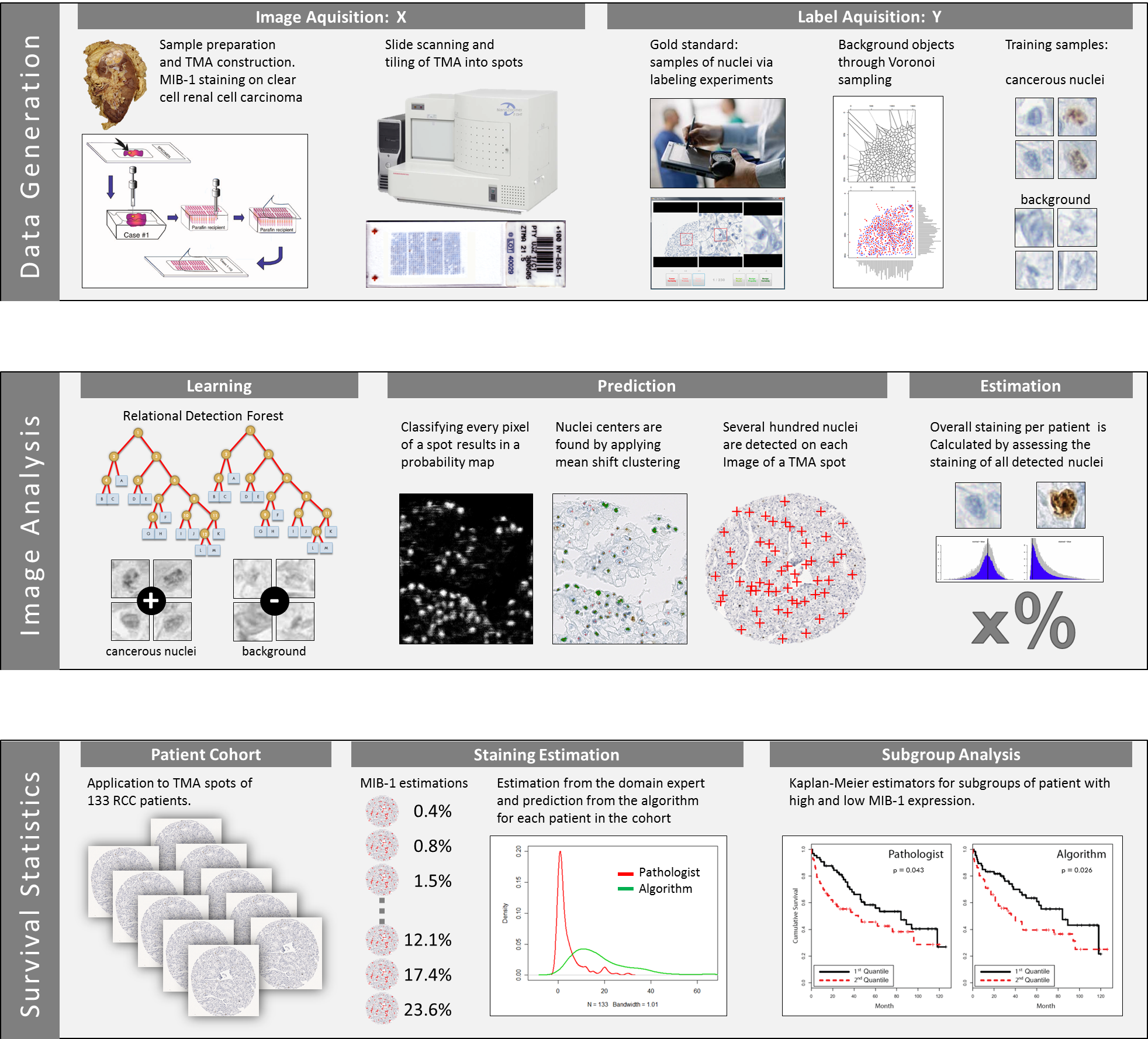}				
  \end{center}			
  \caption{A computational pathology framework for investigating the
  proliferation marker MIB-1 in clear cell renal cell carcinoma.
  Following the definition in Section \ref{ssec:definition} the
  framework consists of three parts: (i) The covariate data $X$
  existing of images of TMA spots was generated in a trial at 
  the University Hospital Zurich. Extensive labeling experiments
  were conducted to generate a gold standard comprising cancerous
  cell nuclei and background samples.
  (ii) Image analysis consisted of learning a relational detection
  forest (RDF) and conducting mean shift clustering for nuclei 
  detection. Subsequently, the staining of detected nuclei was
  determined based on their color histograms.
  (iii) Using this system, TMA spots of $133$ RCC patients were 
  analysed. Finnaly, the subgroup of patients with high 
  expression of the proliferation marker was compared to
  the group with low expression using the Kaplen-Meier estimator.
    \label{fig:comppathrcc}}
\end{figure*}
			
Figure \ref{fig:comppathrcc} depicts a schematic overview of the
project subdivided into the three main parts which are discussed
in the following:

\subsection{Data Generation}
The data generation process consists of acquiring images of the TMA 
spots representing the covariates $X$ in the statistical model and
the target variable $Y$ which comprises detection and classification
labels for nuclei.

The tissue microarray block was generated in a trial at the 
University Hospital Zurich. TMA slides were immunohistochemically 
stained with the MIB-1 (Ki-67) antigen and scanned on a Nanozoomer 
C9600 virtual slide light microscope scanner from HAMAMATSU. 
The magnification of $40\times$ resulted in a per pixel 
resolution of $0.23\mu m$. The tissue microarry was tiled into single 
spots of size $3000\times3000$ pixel, representing one patient each.

Various strategies can be devised to estimate the progression status
of cancerous tissue: (i) we could first detect cell nuclei and then
classify the detected nuclei as cancerous or benign
\cite{FuchsDAGM2008}; (ii) the nucleus detection phase could be merged
with the malignant/benign classification to simultaneously train a
sliding window detector for cancerous nuclei only.
To this end samples of cancerous nuclei were collected using the
labeling experiments described in Section \ref{ssec:labelingex}.
Voronoi Sampling \cite{FuchsISVC2009} was used to generate a set of
negative background patches which are spatially well distributed in
the training images. Hence a Voronoi tessellation is created based on
the locations of the positive samples and background patches are
sampled at the vertices of the Voronoi diagram.  In contrast to
uniform rejection sampling, using a tessellation has the advantage
that the negative samples are concentrated on the area of tissue close
to the nulei and few samples are spent on the homogeneous
background. (The algorithm should not be confused with Lloyd's
algorithm \cite{Lloyd82} which is also known as Voronoi iteration.)
The result of the data generation process is a labeled set of
image patches of size $65 \times 65$ pixel.

\subsection{Image Analysis}
The image analysis part of the pipeline consists of learning
a relational detection forest \cite{FuchsISVC2009} based on
the samples extracted in the previous step. To guarantee
illumination invariance, the feature basis described in 
Section \ref{ssec:invariance} is used.

The strong class imbalance in the training set is accounted for by
randomly subsampling the background class for each tree
of the ensemble. The model parameters are adjusted by
optimizing the out of bag (OOB) error \cite{Breiman01}
and they consist of the number of trees, the maximum tree depth
and the number of features sampled at each node in a tree.

For prediction each pixel of a TMA spot is classified
by the relation detection forest. This results in a 
probability map in the size of the image where the gray 
value at each position indicates the probability of being 
the location of a cancerous nucleus.
Finally, weighted mean shift clustering is conducted with
a circular box kernel based on the average radius $r$ of 
the nuclei in the training set. This process yields the final
coordinates of the detected cancerous nuclei.

To differentiate a stained nucleus from a non-stained 
nucleus a simple color model is learned. Based on the 
labeled nuclei, color histograms are generated for both 
classes based on the pixels within a radius $r$.
A test nucleus is then classified based on the 
distance to the centroid histograms of both classes.

The final staining estimation per patient is achieved
by calculating the percentage of stained cancerous
nuclei.

\subsection{Survival Statistics}

The only objective endpoint in the majority of TMA studies is the
prediction of the number of months a patient survived. The experiments
described in Section \ref{ssec:labelingex} document the large disagreement
between pathologists for the estimation of staining. Hence, fitting an
algorithm to the estimates of a single pathologist or to a consensus
voting of a commitee of pathologist is not desirable.
				
To this end the proposed computational pathology framework is
validated against the right censored clinical survival data of the
$133$ ccRCC patients. In addition these results were compared to the
estimations of an expert pathologist specialized on renal cell
carcinoma. He analyzed all spots in an exceptional thorough manner
which required him more than two hours. This time consuming annotation
exceeds the standard clinical practice significantly by a factor of
$10-20$ and, therefore, the results can be viewed as an excellent
human estimate for this dataset.

\begin{figure}[htbp]
  \begin{center}					
    \includegraphics[width=1\linewidth]{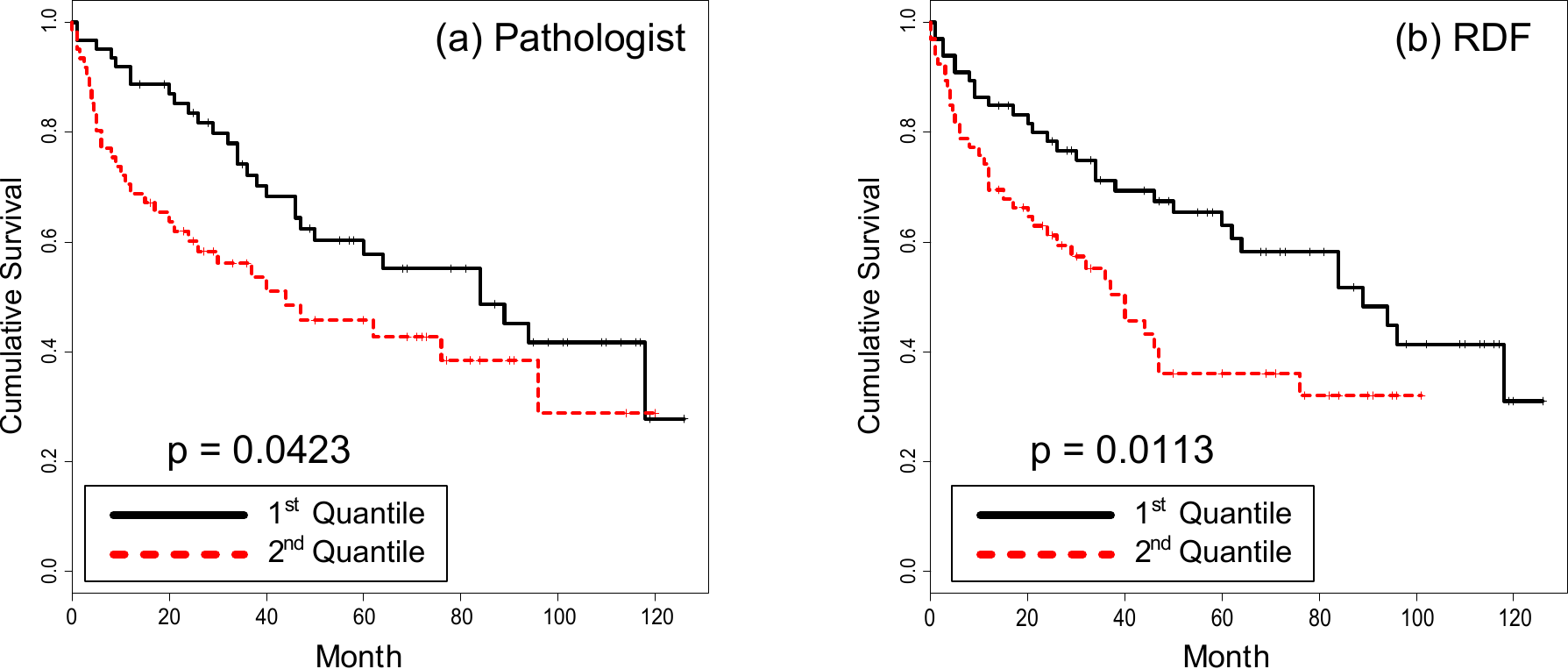}				
  \end{center}			
  \caption{Kaplan-Meier estimators show significantly different
    survival times for renal cell carcinoma patients with high and low
    proliferating tumors. Compared to the manual estimation from the
    pathologist (a) ($p=0.04$), the fully automatic estimation from
    the algorithm (b) compares favorable ($p=0.01$) in terms of
    survival differences (log rank test) for the partitioning of
    patients into two groups of equal size \cite{FuchsISVC2009}.
    \label{fig:surv}}
\end{figure}

Figure \ref{fig:surv} shows Kaplan-Meier plots of the estimated
cumulative survival for the pathologist and the computational
pathology framework. The farther the survival estimates of the two
groups are separated the better the estimation. Quantifying this
difference with a log-rank test shows that the proposed framework
performs favorable ($p=0.0113$) to the trained pathologist
($p=0.0423$) and it can differentiate between the survival
expectancy of the two groups of patients.

\subsection{Project Conclusion}
The presented computational pathology framework can be characterized
by the following properties: (i) \textbf{Simplicity:} It can be used
in a plug-and-play fashion
to train object detectors in near real time for large variety of
tasks. (ii) \textbf{Novel Feature Basis:} The introduced relational
features are able to capture shape information, they are illumination
invariant and extremely fast to evaluate. (iii)
\textbf{Randomization:} The randomized tree induction algorithm is
able to 
exploit the richness of the intractable large feature space and to
take advantage of it by increasing diversity of the ensemble.  (iv)
\textbf{Real World Applicability:} The proposed algorithms perform
well not only on renal cancer tissue but also in fluorescent imaging
of pancreatic islets \cite{FlorosMICCAI09} and in quantifying staining
in murine samples \cite{Bettermann10}.

\section{Future Directions}

\subsection{Histopathological Imaging}
One promising research direction in medical image analysis points to
online learning and interactive learning of computer vision
models. Not only covers histopathology a broad and heterogeneous field but
new biomarkers, antibodies and stainings are developed on a daily
basis. To this end, real world applications have to 
quickly adapt to changing tissue types and staining modalities. Domain
experts should be able to train these models in an interactive fashion
to accustom novel data. For example, a classifier for object detection
can be trained by clicking on novel objects or correcting for false
detections.

A necessary prerequisite for research in computational pathology
proved to be the scanning of whole slides and TMAs. \cite{Huisman10}
describe a fully digital pathology slide archive which has been
assembled 
by high-volume tissue slide scanning.  The Peta bytes of histological
data which
will be available in the near future pose also a number of software
engineering challenges, including
distributed processing of whole slides and TMAs in clusters or the
cloud, multiprocessor and multicore implementations of analysis
algorithms and facilitating real time image processing on GPUs.

\subsection{Clinical Application and Decision Support}

In today's patient care we observe the interesting trend 
to integrate pathological diagnoses in web based patient files.
Avatar based visualization proved to be useful not only for medical
experts but also for a new generation of patients who are better
informed and demand online updated and appropriately visualized
informations about their own disease state and treatment procedures.

Furthermore this approach can be extended for decision support by
statistical models which are able to utilize this unified view of
patients incorporating data from a large variety of clinical sources,
e.g. pathology, cytology, radiology, etc.

\subsection{Pathology@home}
Real-time, in vivo cancer detection on cellular level appears as a futuristic
dream in patient care but could be a reality in a few years.
\cite{Shin10} constructed a fiber-optic fluorescence microscope using
a consumer-grade camera for in vivo cellular imaging. The fiber-optic
fluorescence microscope includes an LED light, an objective lens, a
fiber-optic bundle, and a consumer-grade DSLR. The system was used to
image an oral cancer cell line, a human tissue specimen and the oral
mucosa of a healthy human subject in vivo, following topical
application of $0.01\%$ proflavine.  The fiber-optic microscope
resolved individual nuclei in all specimens and tissues imaged. This
capability allowed qualitative and quantitative differences between
normal and precancerous or cancerous tissues to be identified. In
combination with a computational pathology framework, this technique
would allow the real time classification of cancerous cells in
epithelial tissues.  Such a portable and inexpensive system is
especially interesting for patient care in low-resource settings 
like the developing world.

Constructing a microscope for mobile phones defines the future of
patient care in remote sites with centralized analysis support. 
\cite{Breslauer09} built a mobile phone-mounted light microscope and
demonstrated its potential for clinical use by imaging sickle and
P. falciparum-infected red blood cells in brightfield and
M. tuberculosis-infected sputum samples in fluorescence with LED
excitation. In all cases the resolution exceeded the critical level
that is necessary to detect blood cell and microorganism morphology.
This concept could provide an interesting tool for disease diagnosis
and screening, especially in the developing world and rural areas
where laboratory facilities are scarce but mobile phone infrastructure
is available.

\subsection{Standards and Exchange Formats}
One of the major obstacles for wide spread use of computational
pathology is the absence of generally agreed upon standards and
exchange formats.  This deficit not only handicaps slide processing
management and whole slide digital imaging \cite{Daniel09}, but it
also extends to statistical models and analysis software.
Standardized exchange formats would support project specific
combinations of object detectors, staining estimation algorithms and
medical statistics.  It would be very much desirable if at least the
research community would agree on a few simple interfaces for data and
model exchange.

\subsection{Further Reading}
All links and references presented in this review together with
software, statistical models and a blog about the topic are
available from \href{www.computational-pathology.org}{www.computational-pathology.org}.

\section*{Acknowledgments}
The authors wish to thank Holger Moch and Peter Schraml for their help 
in conducting the RCC TMA project, Peter Wild and Peter Bode for annotating the
medical data and Monika Bieri and Norbert Wey for scanning and tiling
the TMA slides. Special thanks also to Volker Roth and Sudhir Raman for
valuable discussions. We also acknowledge financial support from the FET
program within the EU FP7, under the SIMBAD project (Contract
213250).

\bibliographystyle{elsarticle-num}
\bibliography{main}

\end{document}